%% file: acl_latex.tex
\documentclass[11pt]{article}

\usepackage[final]{acl}

\usepackage{times}
\usepackage{latexsym}

\usepackage[T1]{fontenc}

\usepackage[utf8]{inputenc}

\usepackage{microtype}
\usepackage{inconsolata}

\usepackage{graphicx}

\usepackage{xcolor}
\usepackage{changepage} 
\usepackage{float}
\usepackage{caption}

\usepackage{subcaption}
\usepackage{subfiles}
\usepackage{enumerate}
\usepackage{xurl}   %
\usepackage{hyperref}
\usepackage{tikz}
\usetikzlibrary{arrows.meta,positioning,shapes.misc}

\newcommand{\stagehead}[2]{%
  \begin{minipage}{\linewidth}\centering
    \textbf{#1}\\ \textbf{#2}%
  \end{minipage}%
}

\usepackage[most]{tcolorbox}
\usepackage{dashrule} %

\usepackage{amsmath}
\usepackage{amsfonts}
\usepackage{amssymb}
\usepackage{bbm}
\usepackage{nicefrac}

\usepackage{listings}
\usepackage{multirow}
\usepackage{adjustbox}

\usepackage{booktabs,tabularx,siunitx,array}

\newcolumntype{L}{>{\raggedright\arraybackslash}p{0.26\linewidth}}
\newcolumntype{C}{>{\centering\arraybackslash}p{0.06\linewidth}}

\usepackage{makecell}

\newcommand{\updiff}[1]{\textcolor{green!60!black}{$\uparrow\,$#1}}
\newcommand{\downdiff}[1]{\textcolor{red!60!black}{$\downarrow\,$#1}}

\DeclareFontFamily{U}{stix2bb}{}
\DeclareFontShape{U}{stix2bb}{m}{n} {<-> stix2-mathbb}{}

\NewDocumentCommand{\indicator}{}{\text{\usefont{U}{stix2bb}{m}{n}1}}

\title{Multimodal Conversation Structure Understanding}

\author{%
  Kent K. Chang \qquad Mackenzie Hanh Cramer \qquad Anna Ho\\ 
  {\bf Ti Ti Nguyen} \qquad {\bf Yilin Yuan} \qquad {\bf David Bamman} \\[\medskipamount]
  {\normalfont University of California, Berkeley}\\
  \texttt{kentkchang@berkeley.edu} \\
}

\begin{document}
\maketitle
\begin{abstract}
While multimodal large language models (LLMs) excel at dialogue, whether they can adequately parse the structure of conversation---conversational roles and threading---remains underexplored. In this work, we introduce a suite of tasks and release TV-MMPC, a new annotated dataset, for multimodal conversation structure understanding. Our evaluation reveals that while all multimodal LLMs outperform our heuristic baseline, even the best-performing model we consider experiences a substantial drop in performance when character identities of the conversation are anonymized. Beyond evaluation, we carry out a sociolinguistic analysis of 350,842 utterances in TVQA. We find that while female characters initiate conversations at rates in proportion to their speaking time, they are 1.2 times more likely than men to be cast as an addressee or side-participant, and the presence of side-participants shifts the conversational register from personal to social.
\end{abstract}

\section{Introduction}
\input{01_intro}

\section{Data}\label{sec:dataset}
\input{02_dataset}

\section{Conversation structure understanding}
\input{03_task}

\section{Experiments}\label{sec:experiments}
\input{04_experiments}

\section{Analysis}\label{sec:analysis}
\input{05_analysis}

\section{Conclusion}
\input{06_conclusion}

\section*{Limitations}

This work introduces a complex set of tasks that required developing detailed annotation guidelines and extensive annotators, and ensuring quality entails a time-consuming and labor-intensive process.
We leave large-scale crowdsourcing or other approaches to scale up annotation, which this work enables along with the annotation guidelines, for future work.
To maximize representativeness, we sample clips from TVQA entirely at random. 
This is not exhaustive of the original dataset and may overlook certain conversational patterns.

While the shows in our data feature diverse casts, the conversational dynamics are scripted and directed primarily for American audiences, which inherently privileges Western communicative norms. 
Cues discussed in this work (e.g., direct eye contact to determine addressees) are  also signals that are culturally specific.

We rely extensively on metadata from TMDb and IMDb for actor and character identification; while we believe they are generally reliable and accurate, these sources can include uncredited characters or omit minor roles.
The effectiveness of our TVQA processing pipeline is limited by the capabilities of ASR and face recognition models, which is beyond the scope of this work.

\section*{Ethical considerations}

The dataset presented in this paper is derived from the existing TVQA datasetTVQA~\cite{Lei2018-ri,Lei2020-lq}, which consist of video clips from TV shows and manually created QA pairs. Since the source dataset has already undergone prior curation and is widely used in the research community without known misuse concerns, we believe it poses low risk for dual-use or misuse. No personally identifiable information or sensitive private data is included in either the original or our derived dataset.

\section*{Acknowledgments}

The research reported in this article was supported by resources provided by the Google Gemma Academic Program.

\bibliography{conversation,researchstatements}

\appendix

\input{08_appendices}
\input{09_annotation_guidelines}

\end{document}

%% file: 01_intro.tex
\begin{figure*}[!htbp]
  \centering
  \includegraphics[width=0.85\textwidth]{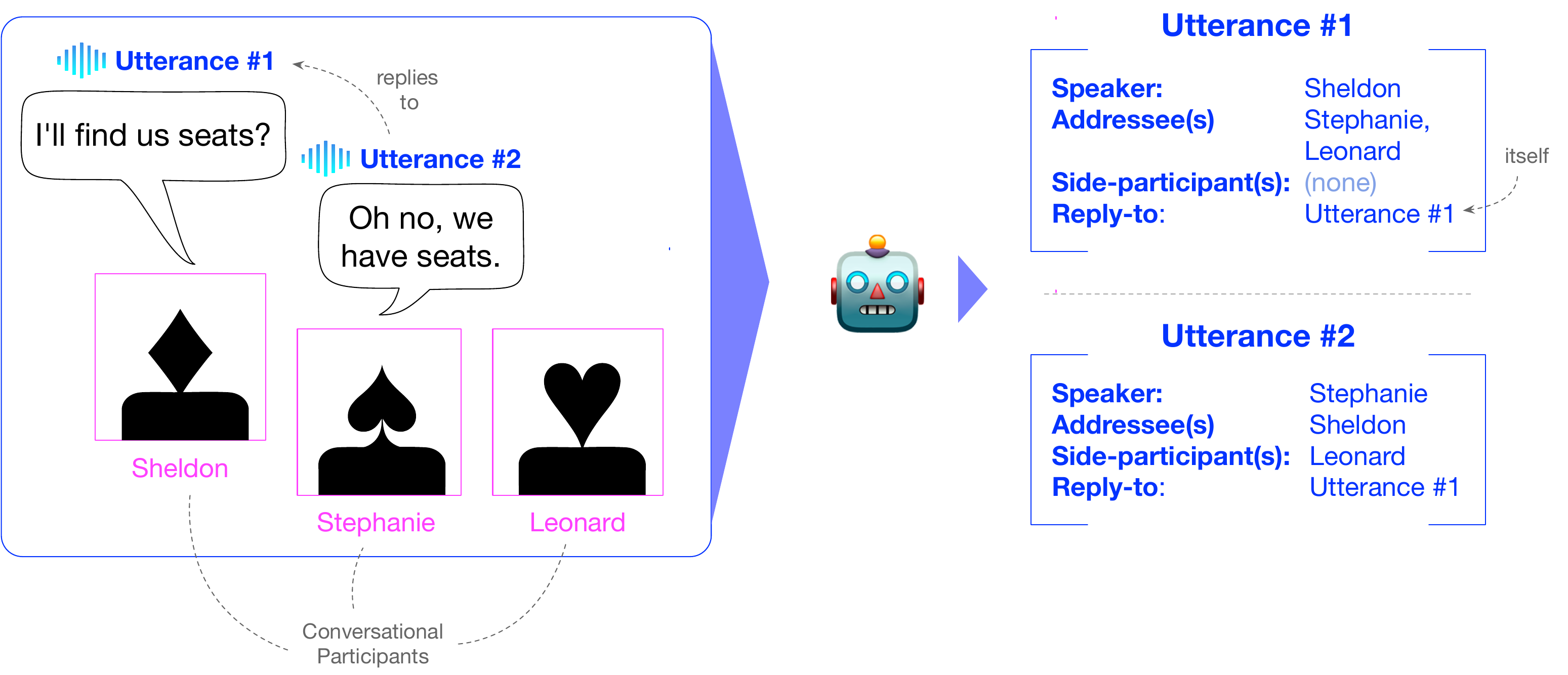}
  \caption{Our proposed structured prediction task for multimodal conversation structure understanding. Grounded in sociolinguistic and conversation analysis~\citep{Goffman1981-fq,Goffman1983-oo,Ng1993-na,Goodwin1981-zs,Clark1982-am}, the task requires predicting, for each utterance in the given clip: the speaker, addressee(s), side-participants, and the utterance it replies to. The example, taken from \textit{The Big Bang Theory} in TVQA~\cite{Lei2018-ri,Lei2020-lq}, illustrates our unified formulation, which treats conversational role attribution and conversation disentanglement as complementary subtasks for modeling the interactional dynamics of dialogue. Further analysis of this example can be found in~Appendix~\S\ref{sec:anno_examples}.}\label{fig:illu}
\end{figure*}

Multi-party conversation understanding involves identifying and structuring interactions between multiple speakers and recipients in a face-to-face setting~\cite{Akhtiamov2019-tw,Lerner2022-mq,Korbar2024-zs}.
Such conversations typically involve complex floor-claiming~\cite{Allan2002-mu,Elsner2008-np,Elsner2010-wi,Wang2020-pz} and turn-taking patterns~\cite{Sacks1974-cg,Hawes2009-fo,Zhou2018-df} marked by various pragmatic cues, which is a challenging problem in NLP~\cite{Gu2022-rq,Tan2023-dx,Penzo2024-hc}. 
Modeling these interactions---often described in sociolinguistics as ritualized social encounters~\cite{Goffman1981-fq}---further enables us to shed light on power dynamics and reveal implicit social hierarchies in conversation.

While multimodal large language models (LLMs) have shown potential in reasoning over complex videos and dialogues~\cite{Song2023-ht,Teng2023-hd,Zhang2023-sd,Bhattacharyya2023-zr,Sun2024-mc}, whether they can adequately parse the structure of conversation---resolve the reply-to relationship between utterances, or attribute roles like speakers and addressees---remains underexplored~\cite{Le_Minh2018-cv,Inoue2025-on}.
To bridge this gap:%

\begin{itemize}
    \item We draw on sociolinguistics~\citep{Goffman1981-fq,Goffman1983-oo,Goodwin1981-zs,Clark1982-am,Ng1993-na} to devise a framework and introduce a suite of tasks (Fig.~\ref{fig:illu}) for capturing the interactional patterns of interlocutors in the multimodal context. %
    \item To support these tasks, we build on TVQA~\cite{Lei2018-ri,Lei2020-lq} and release TV-MMPC, a human-annotated dataset of 4,378 annotations for speakers and reply-to relationship between utterances, 5,599 addressees, and 3,412 side-participants, available under a CC BY-NC 4.0 license.\footnote{Code is available at \url{https://github.com/kentchang/tv-mmpc}; annotations can be downloaded at~\url{https://doi.org/10.7910/DVN/4KUKUL}.}
    \item We find that while the best audio-visual model outperforms others, its performance in speaker and addressee recognition drops significantly when participants are anonymized, casting doubt on model reliability.
    \item In our analysis of 350,842 utterances in TVQA, we find that while female characters initiate conversations at rates proportional to their speaking time, they are more likely to be cast as listeners than speakers, and the presence of side-participants shifts the conversational register from personal to social. This demonstrates the utility of this work in characterizing cultural representation on screen. 
\end{itemize}

%% file: 02_dataset.tex
\begin{figure*}
\centering
\sffamily %

\begin{tikzpicture}[
  font=\scriptsize\sffamily,
  node distance=6mm and 5mm,
  >={Latex[length=1.8mm,width=1.2mm]},
  stage/.style={
    draw, rounded corners=4pt, %
    fill=black!2,
    inner sep=5pt,
    align=left,
    text width=0.19\textwidth
  },
  flow/.style={thick, -{Latex[length=1.8mm,width=1.2mm]}}
]
\node[stage] (s1) {%
  {\stagehead{Stage 1:}{Raw TVQA Data}}\\[4pt]
  60--90s clips reconstructed from frames, audio, and subtitles
};

\node[stage, right=5mm of s1] (s2) {%
  {\stagehead{Stage 2:}{Automated Processing}}\\[4pt]
  Audio re-transcription; speaker label inference and standardization
};

\node[stage, right=5mm of s2] (s3) {%
  {\stagehead{Stage 3:}{Human Annotation}}\\[4pt]
  Verify speakers; annotate addressee, side-participant, reply-to
};

\node[stage, right=5mm of s3] (s4) {%
  {\stagehead{Stage 4:}{TV\textendash MMPC Dataset}}\\[4pt]
  \textbf{4,378} speaker \& reply-to, \textbf{5,599} addressees, \textbf{3,412} side-participants
};

\draw[flow] (s1) -- (s2);
\draw[flow] (s2) -- (s3);
\draw[flow] (s3) -- (s4);
\end{tikzpicture}

\vspace{2pt}
\noindent\hdashrule{\textwidth}{0.35pt}{0.9mm 0.45mm}
\vspace{2pt}

\noindent
\begin{minipage}{0.49\linewidth}
\begin{tcolorbox}[
  enhanced,
  colback=black!2,
  colframe=black!35,
  boxrule=0.6pt,
  arc=3pt,
  title filled,
  colbacktitle=black!85,
  coltitle=white,
  fonttitle=\sffamily\bfseries\scriptsize,
  halign title=left,  
  title={Stage 2: Pre-processing Example},
  boxsep=3pt,
  left=3pt, right=3pt, top=3pt, bottom=3pt,
  toptitle=2pt, bottomtitle=2pt
]
\sffamily\footnotesize

{\scriptsize Before (subtitles):}\par
\vspace{4pt}
\ttfamily\scriptsize
\hspace*{0.5em} 1\\
\hspace*{0.5em} 00{:}00{:}00,677 --> 00{:}00{:}02,257\\
\hspace*{0.5em} (Sheldon:)Wait here, I'll find us seats.\\[2pt]
\hspace*{0.5em} 2\\
\hspace*{0.5em} 00{:}00{:}02,429 --> 00{:}00{:}05,889\\
\hspace*{0.5em} - Oh, no. We have seats.\\
\hspace*{0.5em} - Not the right seats.\\

\normalfont\sffamily\footnotesize
{\scriptsize After (processed TSV):}\par
\vspace{4pt}

\begingroup
\setlength{\tabcolsep}{3pt}      %
\renewcommand{\arraystretch}{1.05}

{\ttfamily\scriptsize
\begin{tabularx}{\dimexpr\linewidth-1em\relax}{@{\hspace{0.5em}} r r l X @{\hspace{0.5em}}}
\toprule
\textbf{start} & \textbf{end} & \textbf{speaker} & \textbf{text} \\
\cmidrule(lr){1-4}
0.031 & 0.711 & sheldon cooper     & I’ll find us seats? \\
1.171 & 2.272 & stephanie barnett  & Oh no, we have seats. \\
2.292 & 3.692 & leonard hofstadter & Not the right seats. \\
\bottomrule
\end{tabularx}
}
\endgroup
\end{tcolorbox}
\end{minipage}
\hfill
\begin{minipage}{0.49\linewidth}
\begin{tcolorbox}[
  enhanced,
  colback=black!2,
  colframe=black!35,
  boxrule=0.6pt,
  arc=3pt,
  title filled,
  colbacktitle=black!85,
  coltitle=white,
  fonttitle=\sffamily\bfseries\scriptsize,
  halign title=left,
  title={Stage 3: Annotation Example},
  boxsep=3pt,
  left=3pt, right=3pt, top=3pt, bottom=3pt,
  toptitle=2pt, bottomtitle=2pt
]
\sffamily\footnotesize

{\ttfamily\scriptsize
\{\\
\ \ "line\_idx": 1,\\
\ \ "speaker": "sheldon cooper",\\
\ \ "addressee": ["stephanie barnett", "leonard hofstadter"],\\
\ \ "side\_participant": [],\\
\ \ "reply\_to": 1\\
\},\\
\{\\
\ \ "line\_idx": 2,\\
\ \ "speaker": "stephanie barnett",\\
\ \ "addressee": ["sheldon cooper"],\\
\ \ "side\_participant": ["leonard hofstadter"],\\
\ \ "reply\_to": 1\\
\}, {\tcbox[
  on line,
  colback=black!12,
  colframe=black!35,
  boxrule=0.3pt,
  arc=2pt,
  top=0pt, bottom=0pt, left=2pt, right=2pt
]{\footnotesize\sffamily . . .}}
} 

\end{tcolorbox}
\end{minipage}
\caption{Data creation pipeline. \textit{Top}: four stages from raw TVQA clips to final annotations. \textit{Bottom}: samples from automated preprocessing (Stage~2) and human annotation (Stage~3), drawn from clip ID \texttt{s02e09\_seg02\_clip\_04}.}
\label{fig:tv-mmpc-pipeline}
\end{figure*}

\begin{table}
  {\centering
\begin{adjustbox}{max width=\linewidth}  
  \footnotesize
  \setlength{\tabcolsep}{5pt}
  \begin{tabular}{lrrr}
    \toprule
                 & \textbf{Speakers} & \textbf{Addressees} & \textbf{Side-participants} \\
    \midrule
    \multicolumn{4}{c}{\textit{Per clip}} \\
    \cmidrule(lr){1-4}
    \# Unique     & 4.18    & 4.48       & 2.90 \\
    \# Avg. Total & 21.89   & 28.00      & 17.06 \\
    \addlinespace[2pt]
    \midrule    
    \multicolumn{4}{c}{\textit{Overall}} \\
    \cmidrule(lr){1-4}    
    \# Unique     & 156     & 168        & 121 \\
    \# Total      & 4{,}378 & 5{,}599    & 3{,}412 \\
    \bottomrule 
  \end{tabular}
\end{adjustbox}
  }
  \caption{Summary statistics for human-annotated data of 200 randomly sampled clips from TVQA.}
  \label{tab:annotation_stats}
\end{table}

In constructing our TV-MMPC (\textbf{M}ulti-modal \textbf{M}ulti-\textbf{P}arty \textbf{C}onversation) dataset, we build upon TVQA~\cite{Lei2018-ri,Lei2020-lq}, a large-scale multimodal dataset, widely used for evaluating multimodal models, consisting of 60–90-second video clips from popular TV series.

We devise a data pre-processing pipeline (Fig.~\ref{fig:tv-mmpc-pipeline}, detailed in Appendix \S\ref{sec:preprop}).
For human annotation, we randomly sample 200 clips from the following shows in TVQA (50 clips each): \textit{The Big Bang Theory}, \textit{Friends}, \textit{House M. D.}, and \textit{How I Met Your Mother}, which captures diverse conversational dynamics ranging from domestic sitcoms to procedural dramas.
To facilitate efficient annotation, we stitch the sampled frames with face captions and use the audio to create a rough approximation of the original clip.
Annotations are carried out by co-authors of the paper, who watch the clips on a dedicated annotation interface and annotate conversational roles and reply-to for each utterance, following the guidelines (cf. Appendix~\S\ref{sec:anno_guidelines}).
Statistics for the annotated dataset can be found in Table~\ref{tab:annotation_stats}.

\begin{table}
  {\centering
\begin{adjustbox}{max width=\linewidth}  
  \footnotesize
  \setlength{\tabcolsep}{5pt}
\begin{tabular}{l rrr}
\toprule
& \multicolumn{3}{c}{\textbf{Inter-annotator agreement}} \\
\cmidrule(lr){2-4}
& {Pilot} & {Main} & {($\Delta$)} \\
\midrule
\multicolumn{4}{l}{\textit{Conversational roles}} \\
\cmidrule(lr){1-4}
Speaker (Acc.)
  & 86.20
  & \textcolor{black}{97.58}
  & (\updiff{11.38}) \\
Addressees (Set F\textsubscript{1})
  & 86.07
  & \textcolor{black}{92.52}
  & (\updiff{6.45}) \\
Side‐part. (Set F\textsubscript{1})
  & 82.87
  & \textcolor{black}{85.43}
  & (\updiff{2.56}) \\
\addlinespace[2pt]
\midrule
\multicolumn{4}{l}{\textit{Conversational threads}} \\
\cmidrule(lr){1-4}
Linking (F\textsubscript{1})
  & 86.07
  & \textcolor{black}{97.11}
  & (\updiff{11.03}) \\
1–NVI
  & 83.85
  & \textcolor{black}{92.87}
  & (\updiff{9.01}) \\
1–1
  & 77.78
  & \textcolor{black}{89.25}
  & (\updiff{11.48}) \\
EM F\textsubscript{1}
  & 35.94
  & \textcolor{black}{70.56}
  & (\updiff{34.62}) \\
\bottomrule
\end{tabular}
\end{adjustbox}
  }
\caption{Annotation quality (pilot vs. main round).}
\label{table:annotation_quality}
\end{table}

We report inter-annotator agreement in Table~\ref{table:annotation_quality} based on a pilot round (10 clips, 221 lines) and a main round (12 clips, 268 lines), where we observe a clear improvement.
Agreement is computed as the average of all pairwise comparisons among the four primary annotators, using the same set of evaluation metrics for experiments (described in Appendix \S\ref{sec:eval_metrics}).

%% file: 03_task.tex
Our framework for conversation structure understanding is grounded in conversation analysis:\footnote{For additional related work, see Appendix~\S\ref{sec:related_work}.}~
we view dialogue not merely as an information exchange, but as 
a ritualized social encounter~\cite{Goffman1981-fq} governed by norms of turn-taking and floor-claiming, which are often mediated by non-verbal cues like gaze and posture. 
To model such conversational dynamics, we introduce the following complementary tasks:

\paragraph{Conversational role attribution.} We draw on the standard tasks of speaker and addressee recognition~\cite{Jovanovic2004-ic,Ouchi2016-wz} and describe a more general task of conversational role attribution.
According to~\citet{Goodwin1981-zs}, conversational roles are observably cast or projected by the \textit{speaker} $s_t$ at time $t$, regardless of what happens at $t+1$ (e.g., addressee might not hear or understand the speaker, and that does not impact our annotation).
We revise the taxonomy proposed in~\citet{Clark1982-am,Clark1987-pn} by organizing participant roles along three binary dimensions, summarized in Table~\ref{tab:participant_roles}: whether a participant is addressed by the speaker, ratified as a member of the conversational group, and known to be attending or perceiving the utterance. 
This reformulation clarifies distinctions among roles and highlights their dependency: 
\textit{addressees} are explicitly addressed, ratified as participants, and known to be listening; 
\textit{side-participants} are not addressed but still ratified and perceptually engaged. 
\textit{Bystanders}, by contrast, are neither addressed nor ratified, and may not be known to be present, which, unlike in~\citet{Clark1987-pn}, includes \textit{overhearers}.%

\begin{table}%
\centering 
\begin{adjustbox}{max width=\linewidth}  
{
\footnotesize
\begin{tabular}{lccc}
\toprule 
\textbf{Role} & \textbf{addressed} & \textbf{ratified} & \textbf{known} \\
\midrule 
Addressee         & $+$ & $+$ & $+$ \\
Side-participants & $-$ & $+$ & $+$ \\
Bystanders        & $-$ & $-$ & $\pm$ \\
\bottomrule 
\end{tabular}}
\end{adjustbox}
\caption{Participant role matrix.} 
\label{tab:participant_roles} 
\end{table}

\paragraph{Conversation disentanglement.} Here, we want to resolve the conversation structure and detect the start of individual sub-conversations or \textit{conversational threads}~\cite{Elsner2010-wi,Gu2022-rq}.
We follow the formulation in~\citet{Chang2023-bb}: given an utterance of interest (UOI) $u_t$, the task is to identify its \textit{parent utterance} $u_p$, which is the utterance that $u_t$ directly replies to: $u_t \rightarrow u_p$. 
In other words, we find the most relevant preceding utterance in the conversational thread; if the utterance marks the start of a new thread, we say it replies to \textit{itself}.
The thread is the transitive closure of such pairwise links.%

Formally, given an input video clip $V$, we first extract the sequence of utterances $\mathcal{U} = \{u_1, u_2, \dots, u_{|\mathcal{U}|}\}$, where each utterance $u_i \in \mathcal{U}$ corresponds to a segment of the video defined by a start time $t_{s, i}$ and an end time $t_{e, i}$, $t_{e, i} \leq t_{s, i+1}$. 
We also identify the set of participants $\mathcal{P} = \{p_1, p_2, \dots, p_{|\mathcal{P}|}\}$ appearing in the video, which are derived from face recognition or other identity tracking methods applied to $V$.
For each UOI, the goal is to predict its speaker, addressees, side-participants, and the utterance to which it directly replies. 
A multi-party conversation structure solver $\mathcal{G}$ maps the current utterance $u_i$ and relevant context to these structural elements:
\begin{equation}
\mathcal{G}: (u_i, \mathcal{U}, \mathcal{P}) \mapsto \{u_p, S_i, \mathcal{A}_i, \mathcal{E}_i\},
\end{equation}
where \textit{reply-to utterance} $u_p \in \{u_1, \dots, u_{i-1}\}$ is the parent utterance that $u_i$ directly replies to, identified by a reply-to function $R: u_i \mapsto u_p$ such that $p < i$. 
If $u_i$ initiates a new thread, $u_p = u_i$; \textit{speaker} $S_i \in \mathcal{P}$ is the participant who produced utterance $u_i$; \textit{addressees} $\mathcal{A}_i \subseteq \mathcal{P}$ is the set of participants primarily addressed by $S_i$ during $u_i$; \textit{side-participants} $\mathcal{E}_i \subseteq \mathcal{P}$ refers to the set of participants actively listening or engaged in the conversation involving $u_i$, but not directly addressed by $S_i$; note that $\mathcal{A}_t \cap \mathcal{E}_t = \varnothing$. 
In this formulation, the \textit{bystanders} are participants who are neither addressees nor side-participants, but nevertheless belong to the same thread and can thus be determined heuristically.

\paragraph{Example.} Consider this scene from \textit{The Big Bang Theory} (season 10, episode 13, clip 6). 
It features four characters in the apartment living room: Leonard, Sheldon, Penny, and Amy. Penny and Amy are preparing to leave for a trip while Leonard and Sheldon are seeing them off.
Our example annotations (see Table~\ref{tab:example}) for this exchange highlight the dynamic nature of conversational interactions.

\begin{table}%
\begin{adjustbox}{max width=\linewidth}
\setlength{\tabcolsep}{4pt}
\begin{tabular}{r l l l l l}
\toprule
\textbf{line} & \textbf{text} & \textbf{speaker} & \textbf{addr.} & \textbf{side-part.} & \textbf{reply-to} \\
\midrule
11 & Have fun.        & Leonard & Penny   & \makecell[l]{Sheldon,\\Amy}      & \makecell[l]{11\\(\texttt{SELF})} \\
12 & Thanks.          & Penny   & Leonard & \makecell[l]{Sheldon,\\Amy}      & 11 \\
13 & Are you ready?   & Penny   & Amy     & \makecell[l]{Sheldon,\\Leonard}  & \makecell[l]{13\\(\texttt{SELF})} \\
14 & Uh-huh.          & Amy     & Penny   & \makecell[l]{Sheldon,\\Leonard}  & 13 \\
\bottomrule
\end{tabular}
\end{adjustbox}
\caption{Example of conversational roles and threads.} 
\label{tab:example} 
\end{table}

In terms of \textit{conversational roles}, in Line 11, Leonard is the \textit{speaker}. He directs his utterance to Penny, making her the \textit{addressee}. Sheldon and Amy, who are present and part of the ratified group but not directly spoken to, are classified as \textit{side-participants}.
The most critical moment for role attribution occurs in Line 13: after replying to Leonard, Penny becomes the speaker and pivots her \textit{attention}:  through non-verbal cues like gaze and turning her body, she selects Amy as the new addressee. This action simultaneously relegates Leonard and Sheldon to the role of side-participants for this new exchange.

The reply-to column in the table captures two distinct, interleaved \textit{conversational threads} within this short exchange:
In Thread 1 (ll. 11–12), Leonard’s ``Have fun'' is marked \texttt{SELF} because it initiates a new social action (i.e., a parting pleasantry) rather than responding to prior dialogue. 
Penny’s response in Line 12, ``Thanks,'' directly replies to Leonard's utterance, and is therefore linked to its parent, Line 11. This completes the first thread.
In Thread 2 (ll. 13–14), Penny’s question, “Are you ready?”, is also marked \texttt{SELF}. 
Although it follows her reply to Leonard, it does not respond to it. 
Instead, it starts a new sub-conversation focused on the logistics of departure, shifting the group's attention. Amy’s “Uh-huh” in Line 14 is a direct reply to this question, linking to its parent, Line 13, and completing the second thread.

Further details and examples can be found in the annotation guidelines in~Appendix \S\ref{sec:anno_examples}.

%% file: 04_experiments.tex
In this section, we evaluate six popular audio-visual models and vision--language models on TV-MMPC (\S\ref{sec:zero-shot}), using the standard set of metrics (described in Appendix~\ref{sec:eval_metrics}).
We analyze the impact of different modalities (\S\ref{sec:ablate}), explore resource-efficient approaches for task adaptation (\S\ref{sec:lora}), and discuss the effects of anonymization (\S\ref{sec:anon}). 

\subsection{Heuristic baseline}

To establish a baseline, we use Whisper~\cite{Radford2022-lm} and pyannote~\cite{Bredin2020-ti} to obtain word-level timestamps and speaker labels.
Since pyannote only provides  generic speaker labels, we determine the speaker by aggregating the frequency of each face appearing at the word level, then assigning the speaker role to the face that appears most frequently within each sentence. 
For each utterance $u_i$, we consider faces present in the context window $[i{-}1, i]$. %
Among these, the most frequently occurring face \textit{excluding} the speaker is designated as the \textit{addressee}, while all remaining faces are labeled as \textit{side-participants}. 
For the linking task, the previous utterance is treated as the parent utterance (i.e., $u_i$ always replies to $u_{i-1}$).

\begin{table*}[t!]%
\centering
\begin{adjustbox}{max width=\textwidth}
\centering
\footnotesize
\begin{tabular}{
  l %
  l %
  l %
  l %
  l %
  l %
  l %
  l %
}
\toprule
& \multicolumn{3}{c}{\textbf{Conversational roles}} & \multicolumn{4}{c}{\textbf{Conversational threads}} \\
\cmidrule(lr){2-4} \cmidrule(lr){5-8}
& \textbf{Speaker} & \textbf{Addressees} & \textbf{Side-part.} & \textbf{Linking} & \multicolumn{3}{c}{\textbf{Clustering}} \\
\cmidrule(lr){2-2} \cmidrule(lr){3-3} \cmidrule(lr){4-4} \cmidrule(lr){5-5} \cmidrule(lr){6-8}
   & Acc. & Set F\textsubscript{1} & Set F\textsubscript{1} & F\textsubscript{1} & 1--NVI & 1--1 & EM F\textsubscript{1} \\
\midrule
\makecell{Heuristic baseline} 
& \makecell{34.67 \\ {\scriptsize [29.13--40.29]}}  %
& \makecell{19.49 \\ {\scriptsize [16.15--23.07]}}  %
& \makecell{36.98 \\ {\scriptsize [31.97--42.14]}}  %
& \makecell{\textbf{92.67} \\ {\scriptsize [90.61--94.60]}}    %
& \makecell{83.34 \\ {\scriptsize [79.24--87.41]}}  %
& \makecell{76.20 \\ {\scriptsize [70.08--82.25]}}  %
& \makecell{31.93 \\ {\scriptsize [19.52--45.92]}}  %
\\[1.5ex]
\midrule
\multicolumn{8}{c}{\textbf{Vision--language models} (image and text)} \\
\midrule
\makecell{LLaMA 4 Scout}
  & \makecell{49.23 \\ {\scriptsize [43.09--55.33]}}   %
  & \makecell{38.47 \\ {\scriptsize [32.94--44.15]}}   %
  & \makecell{42.70 \\ {\scriptsize [35.04--50.76]}}   %
  & \makecell{87.69 \\ {\scriptsize [85.07--90.19]}}   %
  & \makecell{82.60 \\ {\scriptsize [78.51--86.63]}}   %
  & \makecell{76.13 \\ {\scriptsize [70.19--81.96]}}   %
  & \makecell{31.01 \\ {\scriptsize [18.80--43.96]}}   %
\\

\makecell{GPT-4.1 mini} 
& \makecell{55.76 \\ {\scriptsize [49.04--62.32]}} %
& \makecell{46.28 \\ {\scriptsize [40.38--52.21]}} %
& \makecell{53.72 \\ {\scriptsize [46.36--61.32]}} %
& \makecell{81.40 \\ {\scriptsize [78.42--84.25]}} %
& \makecell{78.61 \\ {\scriptsize [74.68--82.50]}} %
& \makecell{75.82 \\ {\scriptsize [70.90--80.64]}} %
& \makecell{21.68 \\ {\scriptsize [11.89--32.49]}} 
\\
\makecell{o4-mini} 
& \makecell{53.39 \\ {\scriptsize [46.72--60.12]}}  %
& \makecell{49.37 \\ {\scriptsize [43.69--55.09]}}  %
& \makecell{56.88 \\ {\scriptsize [48.99--64.71]}}  %
& \makecell{84.65 \\ {\scriptsize [81.50--87.70]}}  %
& \makecell{78.15 \\ {\scriptsize [74.13--82.17]}}  %
& \makecell{74.98 \\ {\scriptsize [69.89--80.02]}}  %
& \makecell{24.87 \\ {\scriptsize [14.79--35.83]}}  %
\\
\makecell{Gemini 2.0 Flash} 
& \makecell{51.78 \\ {\scriptsize [45.45--58.08]}}  %
& \makecell{43.40 \\ {\scriptsize [37.94--48.93]}}  %
& \makecell{53.87 \\ {\scriptsize [46.45--61.44]}}  %
& \makecell{85.46 \\ {\scriptsize [82.67--88.14]}}  %
& \makecell{80.49 \\ {\scriptsize [76.52--84.43]}}  %
& \makecell{77.18 \\ {\scriptsize [72.08--82.21]}}  %
& \makecell{24.37 \\ {\scriptsize [13.65--36.26]}}  %
\\[1.5ex] 
\midrule
\multicolumn{8}{c}{\textbf{Audio-visual LLMs} (video with audio, image, and text)} \\
\midrule
\makecell{Qwen 2.5-Omni 7B}
& \makecell{32.24 \\ {\scriptsize [27.04--37.58]}}  %
& \makecell{23.03 \\ {\scriptsize [18.45--27.91]}}  %
& \makecell{25.94 \\ {\scriptsize [20.22--32.12]}}  %
& \makecell{60.36 \\ {\scriptsize [56.06--64.57]}}  %
& \makecell{66.66 \\ {\scriptsize [62.75--70.67]}}  %
& \makecell{62.19 \\ {\scriptsize [57.51--66.99]}}  %
& \makecell{10.03 \\ {\scriptsize [4.53--16.57]}}  %
\\
\makecell{Reka-Flash}
& \makecell{41.48 \\ {\scriptsize [36.60--46.49]}}  %
& \makecell{35.71 \\ {\scriptsize [30.77--40.69]}}  %
& \makecell{15.56 \\ {\scriptsize [11.99--19.52]}}  %
& \makecell{86.67 \\ {\scriptsize [84.38--88.86]}}  %
& \makecell{81.87 \\ {\scriptsize [77.91--85.73]}}  %
& \makecell{74.41 \\ {\scriptsize [68.40--80.32]}}  %
& \makecell{27.00 \\ {\scriptsize [15.50--39.50]}}  %
\\
\makecell{Gemini 2.0 Flash} 
& \makecell{\textbf{78.60} \\ {\scriptsize [74.21--82.61]}}  %
& \makecell{\textbf{68.11} \\ {\scriptsize [63.29--72.86]}}  %
& \makecell{\textbf{57.68} \\ {\scriptsize [51.20--64.18]}}  %
& \makecell{89.51 \\ {\scriptsize [87.63--91.32]}}  %
& \makecell{\textbf{85.21} \\ {\scriptsize [81.73--88.58]}}  %
& \makecell{\textbf{80.33} \\ {\scriptsize [75.32--85.15]}}  %
& \makecell{\textbf{34.60} \\ {\scriptsize [23.73--46.01]}}  %
\\[1.5ex]
\bottomrule
\end{tabular}
\end{adjustbox}
\caption{LLM zero-shot performance on conversational roles and threads prediction. Metrics are reported with 95\% confidence intervals (CIs) from 10,000 bootstrap resamples.}%
\label{table:zero-shot}
\end{table*}

\subsection{LLM zero-shot performance}\label{sec:zero-shot}

We evaluate the zero-shot performance of two types of models on our tasks: vision--language models (LLaMA 4 Scout, GPT-4.1, o4-mini, and Gemini 2.0 Flash)\footnote{%
\url{https://huggingface.co/meta-llama/Llama-4-Scout-17B-16E},
\url{https://openai.com/index/gpt-4-1/},
\url{https://openai.com/index/introducing-o3-and-o4-mini/},
\url{https://cloud.google.com/vertex-ai/generative-ai/docs/models/gemini/2-0-flash}
} and audio-visual LLMs, including Reka-Flash~\cite{Reka-Team2024-xg}, Qwen 2.5 Omni~\cite{Xu2025-qp}, and Gemini 2.0 Flash with reconstructed clips (cf. \S\ref{sec:dataset}).
We select both open-weight and closed, general-purpose models (i.e., not trained only for captioning or segmentation) to represent distinct approaches to multimodal understanding: vision--language models process sampled frames and text but lack native audio processing, while audio-visual models can process signals from text, audio, and vision.

In our setup, the multimodal LLM $\mathcal{F}$ performs a structured prediction $o_i$ for each $u_i \in \mathcal{U}$:
\begin{equation}
o_i = (o_{s_i},\; o_{a_i},\; o_{e_i},\; o_{r_i}) \sim \mathcal{F}(C_i, \mathcal{P}_{V_i}, Q_{u_i})
\end{equation}
where $C_i$ is the context of $u_i$, 
$\mathcal{P}_{V_i} = \{p_1, p_2, \dots, p_{|\mathcal{P}|}\}$ is the set of identified participants associated with $C_i$, which corresponds to the cast list of the episode from which the clip is taken. 
Participants include the text of character names and, where available, the best face crop between $t_{s, i}$ and $t_{e, i}$.
$Q_{u_i}$ is the question associated with $u_i$: it includes the line index $i$, the start and end timestamps $t_{s, i}$ and $t_{e, i}$, and the question text itself, which prompts the model for conversation structure analysis.
The output ${o}_i$ contains the index of the reply-to utterance $o_{r_i} \leq i$, identities of the predicted speaker $s_i$, addressees $a_i$, side-participants $e_i$; note $o_{a_i}, o_{e_i} \subseteq \mathcal{P}_{V_i}$.
Given the design of models, we need to structure the overall prompt differently for each class of models, particularly what context information $C_i$ to put in the model:

\paragraph{Vision--language models.} The standard setup in video understanding often involves sampling visual frames from a video while discarding its audio. 
Here, we use a modified context $C_i = \mathcal{U}^\star_i$, and $\mathcal{U}_i = \{ (v^\star_j, u_j,\; f_j) \}_{j=1}^{|\mathcal{U}|}$:
$v^\star_i$ is the final frame sampled from $V_i$ in the interval $[t_{s,j}\,, t_{e,j}]$, interleaving with the words of $u_j$; this is further augmented with audio-derived speaker features $f$ extracted using Librosa~\cite{McFee2025-re}: speech rate (slower or faster), pitch range (wide or narrow), pitch mean (lower or higher), and spectral centroid (darker or brighter), which collectively aim to capture the acoustic profile of the speaker.

\paragraph{Audio-visual LLMs.} In this setup, $C_i = (V_i, \mathcal{U}_i)$, where $V_i$ is the input video clip containing utterance $u_i$, i.e., its audio-visual context. It is the same clip that annotators watch (described in \S\ref{sec:dataset}).
The transcription $\mathcal{U}_i$ consists of utterances spoken in $V_i$ ( $u_i\in\mathcal{U}_i$), each represented as a sequence of words with associated start and end timestamps: $\mathcal{U}_i = \left\{ \left( u_j,\; t_{s,j},\; t_{e,j} \right) \right\}_{j=1}^{|\mathcal{U}|}$, where each element corresponds to $j$-th utterance in the clip, and the utterance-level start and end times $t_{s,j}$ and $t_{e,j}$.  %

The results are reported in Table~\ref{table:zero-shot}.\footnote{Additional experimental details are in Appendix \S\ref{sec:experimental_details}.}
All vision--language models we consider perform significantly better than the heuristic baseline across all metrics, other than on the reply-to task.
Another exception is LLaMA 4 Scout, which falls short on side-participant attribution, although it achieves better results on utterance linking than other models, with a statistically significant margin over GPT-4.1 mini.
Among the audio-visual language models we consider, Gemini 2.0 Flash outperforms others by a considerable margin; Qwen 2.5-Omni 7B and Reka-Flash do not show statistically significant improvements over the heuristic baseline on most metrics.
These results suggest the need for further research into multimodal models, particularly in developing methods that more effectively leverage and integrate different input modalities.\footnote{Error analysis for Gemini can be found in Appendix\S\ref{sec:error_analysis}.}

\subsection{Impact of modalities}\label{sec:ablate}

\begin{table}[t!]%
\begin{adjustbox}{max width=\linewidth} 
\centering
\footnotesize
\setlength{\tabcolsep}{2.5pt}
\begin{tabular}{
  l %
  c %
  c %
  c %
  c %
  c %
}
\toprule
& \multicolumn{3}{c}{\textbf{Roles}} & \multicolumn{2}{c}{\textbf{Threads}} \\
\cmidrule(lr){2-4} \cmidrule(lr){5-6}
& \textbf{Spkr.} & \textbf{Addr.} & \textbf{Side-part.} & \textbf{Linking} & \textbf{Clustering} \\
\cmidrule(lr){2-2} \cmidrule(lr){3-3} \cmidrule(lr){4-4} \cmidrule(lr){5-5} \cmidrule(lr){6-6}
\textbf{Modality} & Acc. & Set F\textsubscript{1} & Set F\textsubscript{1} & F\textsubscript{1} & EM F\textsubscript{1} \\
\midrule
\makecell[l]{Text}
  & 36.40 & 36.46 & 40.28
  & 90.67 & 30.73 \\
\makecell[l]{Text + Visual}
  & 51.78 & 43.40 & 53.87
  & 85.46 & 24.37 \\
\makecell[l]{Text + Visual + Audio}
  & 78.60 & 68.11 & 57.68
  & 89.51 & 34.60 \\
\bottomrule
\end{tabular}
\end{adjustbox}

\caption{Modality ablation of Gemini 2.0 Flash.}
\label{table:ablation}
\end{table}

\paragraph{Ablation.} To quantify the specific contribution of individual modalities, we perform an ablation study on our best-performing model, Gemini 2.0 Flash (Table \ref{table:ablation}). 
While adding visual information improves role attribution over text alone, the full audio-visual context is required for high accuracy in speaker and addressee identification.

\paragraph{Error analysis.} To better understand the factors driving performance variation in conversational role attribution, we examine the correlations between interpretable, clip-level features (across textual, acoustic, and visual modalities) and F\textsubscript{1} scores. 
We employ Spearman's $\rho$ to assess monotonic relationships: a positive $\rho$ indicates that higher feature values consistently co-occur with higher performance ranks.

\begin{figure*}[t!]
  \centering
  \includegraphics[width=0.98\textwidth]{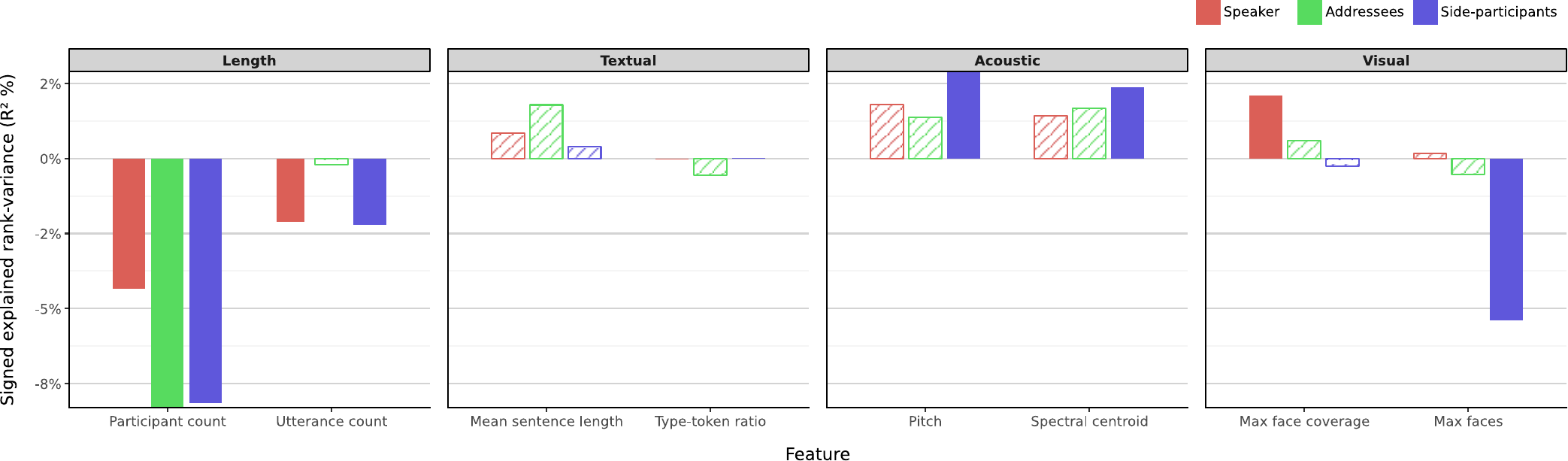}
  \caption{Signed explained rank variance from Spearman's $\rho$ between clip-level features and F\textsubscript{1} scores for individual conversational roles. Bars indicate direction and magnitude of correlation: solid ones are significant ($p<0.05$).}\label{fig:corr}
\end{figure*}

Fig.~\ref{fig:corr} displays the signed explained rank-variance: $R^2 = \text{sign}(\rho)\cdot\rho^2\times100$, where solid bars denote statistical significance ($p<0.05$).
Unsurprisingly, the number of participants has the strongest negative impact, reducing F\textsubscript{1} scores across all roles (3--8\% variance). Clip length shows negligible influence. 
For textual features, we consider mean sentence length (measured by tokens separated by whitespace) and type--token ratio; neither shows significant correlation, suggesting lexical diversity alone does not drive performance.
Acoustic features (pitch, spectral centroid) positively correlate with side-participant attribution ($\rho=+0.17$ and $+0.15$), which indicates that speech clarity aids in identifying peripheral roles.
Visual features reveal role-specific dynamics: max face coverage positively correlates with speaker attribution ($\rho=+0.15$), suggesting that accurate face detection benefits this task, while raw face counts negatively correlate with side-participant detection ($\rho=-0.23$), thereby confirming that visual clutter degrades performance.

Overall, performance relies heavily on acoustic and visual fidelity rather than textual signals.
We suggest that future work look beyond text to ground conversation structure in the multimodal context, and for future audio-visual LLMs, cross-modal cues (e.g., face--participant identity alignment, prosodic salience, etc.) have the potential to enable better parsing of conversation structure.

\subsection{Resource-efficient supervised fine-tuning}\label{sec:lora}

Given the increasing popularity of audio-visual LLMs and the growing number of open-source implementations in this space, we turn to explore task-specific adaptation through supervised fine-tuning (SFT).
In this subsection, we describe our approach using Low-Rank Adaptation (LoRA) to fine-tune Qwen 2.5-Omni 7B within the LLaMA Factory framework~\cite{Zheng2024-da}.
We employ a leave-one-show-out evaluation strategy: training the model on all \textit{but} one show, and testing it on the held-out show.
This process is repeated for each show in the dataset.
The results are reported in Table~\ref{table:zero-shot}, which we compare the performance before and after LoRA SFT on the same TV series.
After fine-tuning, despite the constraint in resources, the model has a statistically significant boost on both addressee and side-participant recognition.

\begin{table}%
\begin{adjustbox}{max width=\linewidth} 
\centering
\footnotesize
\setlength{\tabcolsep}{5pt}
\begin{tabular}{l rrr}
\toprule
& \multicolumn{3}{c}{\textbf{Qwen 2.5-Omni 7B}} \\
\cmidrule(lr){2-4}
& {Zero-shot} & {LoRA SFT} & {($\Delta$)} \\
\midrule
\multicolumn{4}{l}{\textit{Conversational roles}} \\
\cmidrule(lr){1-4}
Speaker (Acc.)
  & 32.24
  & \textcolor{gray!70!black}{41.14}
  & (\updiff{8.90}) \\
Addressees (Set F\textsubscript{1})
  & 23.03
  & \textcolor{gray!70!black}{\textbf{34.90}}
  & (\updiff{\textbf{11.87}}) \\
Side‐part. (Set F\textsubscript{1})
  & 25.94
  & \textcolor{gray!70!black}{\textbf{57.69}}
  & (\updiff{\textbf{31.74}}) \\
\addlinespace[2pt]
\midrule
\multicolumn{4}{l}{\textit{Conversational threads}} \\
\cmidrule(lr){1-4}
Linking (F\textsubscript{1})
  & 60.36
  & \textcolor{gray!70!black}{60.40}
  & (\updiff{0.04}) \\
1–NVI
  & 66.66
  & \textcolor{gray!70!black}{72.82}
  & (\updiff{6.16}) \\
1–1
  & 62.19
  & \textcolor{gray!70!black}{71.20}
  & (\updiff{9.01}) \\
EM F\textsubscript{1}
  & 10.03
  & \textcolor{gray!70!black}{\textbf{32.00}}
  & (\updiff{\textbf{21.97}}) \\
\bottomrule
\end{tabular}
\end{adjustbox}
\caption{Performance of LoRA SFT of Qwen 2.5-Omni 7B compared to zero-shot. Boldface indicates statistically significant (non-overlapping 95\% CIs) differences.}
\label{table:qwen-lora}
\end{table}

\subsection{Effects of anonymization}\label{sec:anon}

\begin{table}[t]
\centering
\begin{adjustbox}{max width=\linewidth}
\footnotesize
\setlength{\tabcolsep}{5pt}
\begin{tabular}{l rrr}
\toprule
& \multicolumn{3}{c}{\textbf{Gemini 2.0 Flash} (audio\mbox{-}visual)} \\
\cmidrule(lr){2-4}
& {Original} & {Anonymized} & {($\Delta$)} \\
\midrule
\multicolumn{4}{l}{\textit{Conversational roles}} \\
\cmidrule(lr){1-4}
Speaker (Acc.)
  & 78.60
  & \textcolor{gray!70!black}{\textbf{13.68}}
  & (\downdiff{\textbf{64.92}}) \\
Addressees (Set F\textsubscript{1})
  & 68.11
  & \textcolor{gray!70!black}{\textbf{15.73}}
  & (\downdiff{\textbf{52.37}}) \\
Side‐part. (Set F\textsubscript{1})
  & 57.68
  & \textcolor{gray!70!black}{46.06}
  & (\downdiff{11.62}) \\
\addlinespace[2pt]
\midrule
\multicolumn{4}{l}{\textit{Conversational threads}} \\
\cmidrule(lr){1-4}
Linking (F\textsubscript{1})
  & 89.51
  & \textcolor{gray!70!black}{88.44}
  & (\downdiff{1.08}) \\
1–NVI
  & 85.21
  & \textcolor{gray!70!black}{84.56}
  & (\downdiff{0.65}) \\
1–1
  & 80.33
  & \textcolor{gray!70!black}{80.14}
  & (\downdiff{0.18}) \\
EM F\textsubscript{1}
  & 34.60
  & \textcolor{gray!70!black}{33.49}
  & (\downdiff{1.11}) \\
\bottomrule
\end{tabular}
\end{adjustbox}
\caption{Impact of anonymization on Gemini (bold indicates non-overlapping 95\% CIs).} %
\label{tab:anonymization}
\end{table}

Prior work has shown that anonymizing speakers can influence downstream performance in text-based dialogue understanding~\cite{Chen2022-fg,Chang2024-yr}. 
Motivated by this, we examine whether anonymizing conversational participants has a similar impact in our multimodal setting. 
To do so, for each clip, we create a mapping based on its individual episode cast list: we assign a random letter to each character within a clip (e.g., ``Sheldon Cooper'' becomes \texttt{Character C}). 
We then apply the same frame captioning procedure described in~\S\ref{sec:dataset}, but we replace those actual character names with their anonymized labels.

The difference in performance is presented in Table~\ref{tab:anonymization}. 
We observe significant drops in the performance for speaker and addressee recognition, by 64.92 and 52.37 points, respectively. 
In light of this, future work might assess the degree to which the performance drop can be attributed to the phenomenon of memorization in the audio-visual LLMs, which is relevant in the context of vision--language models~\cite{Jayaraman2024-kw} and ASR models~\cite{Wang2024-fj}. 
This is more than a straightforward case of test set contamination (where test data overlaps with training data), but a subtler one: models, when solving a task, might leverage their parametric knowledge, among other things, acquired from exposure to cultural artifacts and information about them, potentially across different modalities. 
Such memorization raises further questions about model reliability and cultural representation, which has been explored in the context of literary texts~\cite{Chang2023-oy,Walsh2024-in} and can be relevant to audio-visual LLMs.

%% file: 05_analysis.tex
We posit that a robust structural understanding of conversation serves as the necessary scaffolding for analyzing power dynamics and cultural representation. 
To demonstrate the affordance of this work, we apply our best-performing model to analyze the clips in TVQA we do not annotate, which comprises 350,842 utterances across four TV series, and present three case studies below. %

\subsection{Gender and floor claiming}\label{sec:analy1}

\begin{figure}[t!]
  \centering
  \includegraphics[width=1.02\columnwidth]{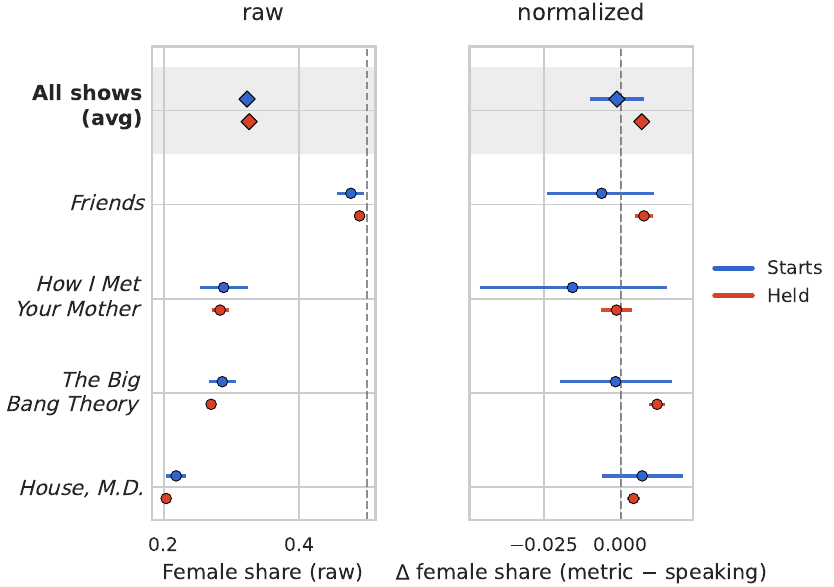}
  \caption{
  Female share of starting vs. holding conversational threads, aggregated by show and overall. \textit{Left}: raw percentage of threads started or held by women. \textit{Right}: normalized difference ($\Delta$) between female share of metric and speaking time within each clip.  Points are bootstrapped means (95\% CIs).
  }
  \label{fig:show_thread_hold}
\end{figure}

How is the act of claiming and holding conversational floor distributed between genders? 
To explore this, we measure two aspects of conversational agency across 9,780 conversation-initiating utterances with known speaker gender: who \textit{starts} new threads mid-clip, and who \textit{holds} the thread by being the recipient of a reply. 
Given the well-documented disparity in speaking time in media~\cite{Lauzen2019-ta,OMeara2016-ek}, we analyze both raw counts and per-clip metrics normalized by lifting each character's share of speaking time.

Fig.~\ref{fig:show_thread_hold} presents these dynamics. In raw terms, the female share of starting threads is low ($31.8\%$), confirming that men dominate this action in absolute numbers. 
However, the right panel (the normalized difference) reveals a more nuanced picture. For thread \textit{starters}, the average delta is not statistically different from zero (mean $\Delta = -0.12\%$, $p=0.79$), indicating that women initiate threads at a rate proportional to their speaking time. 
In contrast, for thread \textit{holders}, the average delta is small but statistically significant and positive (mean $\Delta = +0.69\%$, $p < 0.001$). 
This suggests a dynamic where female characters not only exhibit agency in starting conversations but also receive slightly more conversational uptake in the form of replies than their speaking time would predict.

\subsection{Gender and conversational roles}\label{sec:analy2}

{
\newlength{\imgboxheight}
\setlength{\imgboxheight}{4.5cm} 

\begin{figure*}[t]
  \centering
  \begin{subfigure}[t]{0.342\textwidth}
    \centering
    \subcaption{}
    \begin{minipage}[t][\imgboxheight][t]{\linewidth}
      \includegraphics[width=\linewidth,keepaspectratio]{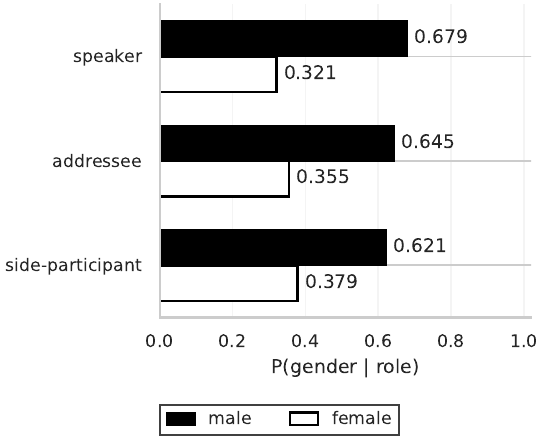}
    \end{minipage}
  \end{subfigure}\quad%
  \begin{subfigure}[t]{0.272\textwidth}
    \centering
    \subcaption{}
    \begin{minipage}[t][\imgboxheight][t]{\linewidth}
      \includegraphics[width=\linewidth,keepaspectratio]{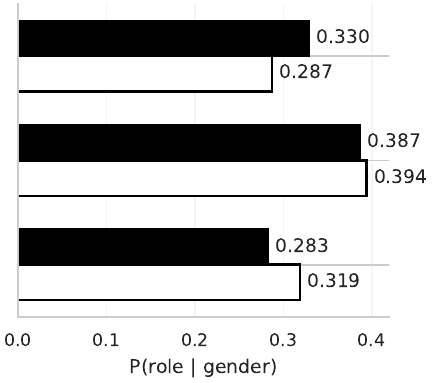}
    \end{minipage}
  \end{subfigure}\quad%
  \begin{subfigure}[t]{0.143\textwidth}
    \centering
    \subcaption{}
    \begin{minipage}[t][\imgboxheight][t]{\linewidth}
      \includegraphics[width=\linewidth,keepaspectratio]{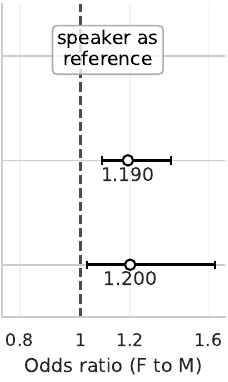}
    \end{minipage}
  \end{subfigure}
  \caption{
  The relationship between gender and conversational roles. (a) $P(\text{gender} | \text{role})$: the probability that a participant in a given role is female or male; (b) $P(\text{role} | \text{gender})$: the probability distribution of roles for each gender; (c) odds ratios from a multinomial regression, with speaker as the reference.
  }
  \label{fig:three-panels-tablelike}
\end{figure*}
}

Beyond who starts or holds a thread, how are men and women positioned within the broader structure of conversations? 
We analyze the distribution of conversational roles across over 929,000 instances in our dataset.
Fig.~\ref{fig:three-panels-tablelike} breaks down this relationship. 
Panel (a), showing $P(\text{gender}|\text{role})$, confirms that men are the numerical majority in every role. 
However, Panel (b), which shows $P(\text{role}|\text{gender})$, reveals a structural disparity:
Men are more likely than women to occupy the \textit{speaker} role ($33.0\%$ of their roles vs. $28.7\%$ for women). 
Conversely, this pattern inverts for listening roles: women are more likely than men to be cast as an \textit{addressee} ($39.4\%$ vs. $38.7\%$) and a \textit{side-participant} ($31.9\%$ vs. $28.3\%$). 
This skewed distribution positions male characters as primary speakers and female characters more frequently as the audience.
Panel (c) formalizes this disparity using a multinomial logistic regression. With speaker as the reference category, the model estimates the log-odds of a person being an addressee or side-participant, controlling for gender and show-level fixed effects:

{\small\begin{align}
\log \frac{P(\text{role}=j)}{P(\text{role}=\text{speaker})} = \beta_{j,0} &+ \beta_{j,\text{female}} \cdot \mathbb{I}(\text{female}) \notag\\
 & + \sum_{s=1}^{S-1} \gamma_{j,s} \cdot \mathbb{I}(\text{show}_s),    
\end{align}}

\noindent where $j \in \{\text{addressee, side-participant}\}$ and $\mathbb{I}(\cdot)$ is an indicator function, and the odds ratio for being female is given by $\exp(\beta_{j,\text{female}})$. 
These odds ratios are $1.19$ for addressee ($p<0.001$, $95\%$ CI $[1.08, 1.39]$) and $1.20$ for side-participant ($p<0.001$, $95\%$ CI $[1.02, 1.64]$). 
This suggests that, controlling for show-specific effects, women are significantly more likely than men to be cast as a listening role than as the primary speaker.

\subsection{Audience design}\label{sec:analy3}

Speakers dynamically adapt their linguistic style based on their audience; to investigate whether such \textit{audience design}~\cite{Bell1984-hf,Giles1991-aa} occurs, we identify distinctive lexical items using a stratified analysis based on the weighted log-odds ratio with an informative Dirichlet prior~\cite{Monroe2017-us}.
First, we place an informative Dirichlet prior on each term $t$, where the prior count $\alpha_t = C^* \cdot p_t$ is weighted by the term's background frequency $p_t$. Second, the prior strength $C^*$ is set via Empirical Bayes calibration: we permute group labels within each show and select the $C^*$ from a grid that results in the null z-scores having a standard deviation closest to 1. 
Finally, with the calibrated prior, we compute a per-show z-score ($\hat{\zeta}_{ts}$) for each term. This score is the log-odds ratio, $\hat{\delta}_{ts}$, scaled by its variance, $\sigma^2(\hat{\delta}_{ts}) \approx (y_{tsa} + \alpha_t)^{-1} + (y_{tsb} + \alpha_t)^{-1}$. These per-show scores are then aggregated across all $k$ shows using equal-weight Stouffer's method to produce a final, overall z-score, $Z_t = \nicefrac{1}{\sqrt{k}} \sum_{s=1}^k \hat{\zeta}_{ts}$.

\begin{table}[t]
\centering
\begin{adjustbox}{max width=\linewidth}
\small %
\setlength{\tabcolsep}{12pt}
\begin{tabular}{l r l r}
\toprule
\multicolumn{4}{c}{\textbf{No side-participants vs. side-participants present}} \\
\midrule
\multicolumn{2}{c}{\textit{more private}} & \multicolumn{2}{c}{\textit{more public}} \\
\cmidrule(lr){1-2} \cmidrule(lr){3-4}
term & score & term & score \\
\midrule
\texttt{i've} & $2.54$ & \texttt{happened} & $-2.41$ \\
\texttt{liver} & $2.18$ & \texttt{ball} & $-2.17$ \\
\texttt{job} & $2.13$ & \texttt{lovely} & $-2.15$ \\
\texttt{dum} & $2.02$ & \texttt{possible} & $-2.02$ \\
\texttt{went} & $2.02$ & \texttt{list} & $-2.00$ \\
\texttt{idiot} & $1.96$ & \texttt{next} & $-1.99$ \\
\texttt{without} & $1.87$ & \texttt{princeton} & $-1.92$ \\
\texttt{straight} & $1.81$  & \texttt{fluid} & $-1.86$ \\
\texttt{hire} & $1.80$ & \texttt{how's} & $-1.82$ \\
\texttt{first} & $1.79$  & \texttt{ice} & $-1.74$ \\
\bottomrule
\end{tabular}
\end{adjustbox}
\caption{Distinctive terms illustrating a shift from private to public conversational register. The table lists the top 10 terms most associated with whether side-participants are present, ranked by their final z-scored log-odds ratios. Positive scores indicate terms characteristic of private, dyadic speech, while negative ones indicate terms characteristic of public, group speech.}
\label{tab:audience_design_register}
\end{table}

Table~\ref{tab:audience_design_register} highlights the results of this analysis by comparing speech in private, dyadic settings (i.e., no side-participants) with speech in more public, group settings (i.e., side-participants present). 
The linguistic patterns reveal a clear shift in register: language in private contexts trends towards the transactional and personal, featuring first-person experiences (\texttt{i've}), direct tasks (\texttt{job}), and even confrontational terms (\texttt{idiot}).
In contrast, language used when side-participants are present is more social and performative, including politeness markers (\texttt{lovely}) and terms that manage group conversation (\texttt{happened}, \texttt{how's}). 
Taken together, this lexical analysis represents an instance of audience design, where speakers would alter their vocabulary in response to their interlocutors.

%% file: 06_conclusion.tex
Conversations are usually structured by roles---who is speaking, who’s being addressed, and who's listening---and unfold in threads that break with changes in speaker floor or topical focus. 
In this work, we address the limited evaluation of multimodal LLMs on conversation structure by introducing a unifying framework for conversational role attribution and thread disentanglement.
To support this work, we present TV-MMPC, a human-annotated dataset built on TVQA, with 4,378 speaker and reply-to annotations, 5,599 addressee labels, and 3,412 side-participant tags.

Our experiments reveal that while Gemini 2.0 Flash outperforms baselines by leveraging audio-visual context, its performance drops significantly under anonymization, opening new avenues for future work.
To demonstrate the affordance of this work, we carry out a sociolinguistic analysis and uncover a communicative hierarchy where female characters are 1.2 times more likely to be cast as listeners and observe that side-participants shift dialogue registers from personal to performative. 

We hope this work can enable future work to improve multimodal LLMs and provide more evidence of how social roles and power dynamics are constructed---and reinforced---in cultural artifacts.
Code is available at \url{https://github.com/kentchang/tv-mmpc}; annotations can be downloaded at~\url{https://doi.org/10.7910/DVN/4KUKUL}.

%% file: 08_appendices.tex
\section{Data pre-processing for annotation}\label{sec:preprop}

To complement Fig.~\ref{fig:tv-mmpc-pipeline}, we include further descriptions of our data pre-processing pipeline below:

\paragraph{Re-transcribe the audio on the sentence-level.} The original subtitles lack clear sentence boundaries, which makes them unsuitable for annotation aimed at understanding conversational structure. 
To address this, we transcribe the raw audio using Whisper~\cite{Radford2022-lm} and align the transcription with the original subtitle data from TVQA.

\paragraph{Infer and standardize speaker labels.} We are interested in \textit{conversational participants}, which, in the context of TV series, are all characters that appear in the clip.
For this reason, we focus on post-processing. 
Some of the subtitles from TVQA contain speaker labels; those labels are usually the first names of the characters in the clip. 
To facilitate downstream face recognition and annotation, we standardize those labels and map them to actor names by querying TMDb\footnote{\url{https://www.themoviedb.org/}} for the cast list of each episode.
This gives us the mapping between the canonical names of the characters and the actors who play them.
With this information, we project TVQA speaker labels where possible (i.e., when the label is present and not unknown for the given sentence) and the algorithm described in~\cite[p. 24]{Huang2020-vc} to align TVQA subtitles and Whisper transcriptions. 
For a second pass, we infer speaker labels from the sampled frames and audio: 
we use the most frequently occurring face during the speech in question,~
and we recognize the face following the pipeline described in~\citet{Bamman2024-qu}, which maps the faces to actor names on IMDb,\footnote{\url{https://www.imdb.com/}}
and the character who dominates the duration of the sentence is treated as the speaker of the sentence.
Those speaker labels are verified during the annotation stage; if no face or actor is detected from the automatic pipeline described above, we manually annotate them too.

\section{Additional related work}\label{sec:related_work}

\paragraph{Multi-party conversation understanding.} Multi-party conversation understanding (i.e., dialogues with more than two participants) can be summarized as solving ``who says what to whom''~\cite{Gu2022-rq}. 
This work tends to consider a limited set of modalities, including text-~\citep{Lu2022-ws,Tan2023-dx,Penzo2024-hc} or audio-only~\citep{Akhtiamov2019-tw,Lerner2022-mq} approaches.

Representative tasks include addressee recognition and conversation disentanglement; both require understanding the relationship between speakers---or more generally, conversational participants---and their utterances. 
They are designed and built upon text-based data, such as chatroom logs~\cite{Kummerfeld2019-li,Gu2021-es}.
Works involving audio-visual data focus on speaker activity, identity, and visual co-occurrence~\cite{Bredin2016-dh,Roth2020-xm,Sharma2023-to} and their long-range dependencies~\cite{Tao2021-ce,Kim2024-tu}, mostly without resolving the semantics of conversation structure, until more recently~\cite{Korbar2024-zs}. 

Text--vision models leverage image captions or scene descriptions to infer conversational roles in the broader video understanding framework, which suggests a focus on reasoning over the visual aspect of people and objects in the frame that ignores speech~\citep{Zellers2019-zs,Lei2020-lq}.
In this work, we believe that multiple modalities are useful in capturing different aspects of the conversation structure, and a robust understanding of conversational dynamics should mirror human understanding of social interactions.

\paragraph{Video understanding.} Video understanding involves reasoning over an extended temporal context that includes a sequence of images and/or audio signals.
It is fundamental to tasks ranging from movie genre classification to scene graph creation~\cite{Huang2020-vc,Vicol2018-ao,Wu2021-hx,Islam2022-vo,Mangalam2023-vz,Zhang2023-sd,Sun2024-mc}.
Such work focuses on multimodal representations of video and parsing interactions between visible objects, exemplified by visual commonsense reasoning (VCR,~\citealp{Zellers2019-zs}) and TVQA~\citep{Lei2018-ri,Lei2020-lq}, require models to infer plausible interactions between characters in an image (or a sequence of frames sampled from videos) and reason over their actions.
However, they typically do not explicitly resolve the underlying conversation structure and often discards speech data (audio or transcription).

Those works converge in recent advances of foundational models: they are natively multimodal~\cite{Gemini-Team2023-oj,OpenAI2024-jl,Reka-Team2024-xg,AI-Meta2025-ph}, general-purpose models that can also reason over videos.
In this work, we assess the capabilities of such models to resolve the conversation structure of such long-form videos.
Related work that bears most resemblance to this study does consider multiple modalities for solving conversation structure (addressee recognition) but involves a controlled setting~\citep{Le_Minh2018-cv,Inoue2025-on}.
This work represents interactions and settings represented on screen, and we discuss the impact of this in~\S\ref{sec:error_analysis}.

\section{Evaluation metrics}\label{sec:eval_metrics}

Following existing work, we use the following metrics to evaluate the performance of models:

\subsection{Conversational roles}

\paragraph{Accuracy.} Speaker identification is evaluated by comparing the predicted speaker label with the true label: $\nicefrac{1}{N} \sum_{i=1}^{N} \indicator(y_i = \hat{y}_i)$, where $y_i$ is the true label, $\hat{y}_i$ is the predicted label.
This is equivalent to precision @1 used in previous work~\cite{Gu2023-if}.

\paragraph{Set-Based F\textsubscript{1}.} Existing work uses F\textsubscript{1} scores for addressee recognition, treating it as a multi-class classification task.
Our setup allows multiple addressees and side-participants, so we adapt multi-label metrics that compare gold (annotated) and predicted sets of participants: $\nicefrac{1}{N} \sum_{i=1}^N \text{F1}_i$, where $\text{Precision}_i = \nicefrac{|\hat{Y}_i \cap Y_i|}{|\hat{Y}_i|}, \text{Recall}_i = \nicefrac{|\hat{Y}_i \cap Y_i|}{|Y_i|}, \text{F1}_i = \nicefrac{2 \cdot \text{Precision}_i \cdot \text{Recall}_i} {\text{Precision}_i + \text{Recall}_i}$, and $Y_i$ is the gold set and $\hat{Y}_i$ is the predicted set for utterance $i$.

\subsection{Conversational thread}

Reply-to involves first identifying the pairwise links between an utterance of interest and its parent utterance (child utterance \textit{replies to} the parent), and secondly turning those links into conversational threads. 
We can see the former as equivalent to binary classification (i.e., whether two utterances exhibit the reply-to relationship) and the second as a clustering problem. 
The former is evaluated using the F\textsubscript{1} score; the latter on the cluster-based metrics below:

\paragraph{Normalized Variation of Information (NVI).} This metric measures the agreement between two sets of clusters. Based on information theory, it asks, ``How much information does the predicted clustering give me about the actual (gold) clustering?'' A higher score indicates that the predicted and gold threads are more structurally similar.
Formally, this metric measures the dissimilarity between clustering assignments $C$ and $C'$, based on entropy and mutual information:
\begin{align}
\text{VI}(C, C') &= H(C) + H(C') - 2I(C, C')\\
\text{NVI}(C, C') &= 1 - \frac{\text{VI}(C, C')}{\log_2 N},
\end{align}
where \( H(C) \) is the entropy of clustering \( C \), \( I(C, C') \) is the mutual information, and \( N \) is the total number of utterances.
We report $100 \times \left( 1 - \text{NVI} \right)$, so larger is better.

\paragraph{One-to-One Overlap (1--1,~\citealp{Elsner2008-np}).} This is a more forgiving metric that rewards identifying the \textit{essence} of each conversational thread. It finds the best possible one-to-one pairing between the predicted and gold clusters to maximize their shared utterances. It doesn't require a perfect match, so it gives credit even if a predicted thread is mostly correct but has a few extra or missing utterances.
Formally, this metric finds a one-to-one mapping between gold and predicted clusters that maximizes their overlap. 
It treats the contingency matrix $M$, where $M_{i,j}$ is the number of shared utterances between gold cluster $i$ and predicted cluster $j$, as a weighted bipartite graph. 
The optimal assignment is solved using a min-cost flow algorithm, which identifies the pairing of clusters that yields the greatest total intersection:
\begin{equation}
\text{One-to-One Overlap} = \frac{\max_{\pi} \sum_{i} M_{i, \pi(i)}}{\sum_{i,j} M_{i,j}},    
\end{equation}
where $\pi$ is a one-to-one mapping from gold to predicted clusters.

\paragraph{Exact Match (EM,~\citealp{Kummerfeld2019-li}).} This is our strictest metric; it measures the perfect recovery of threads. A predicted cluster only gets credit if it is an identical match to a gold-standard cluster (containing the exact same set of utterances). This score reflects the model's ability to precisely identify the complete boundaries of a conversation.
Formally, this metric computes precision, recall, and F\textsubscript{1}-score based on exact identity between individual gold-standard clusters $c_i \in C_{\text{gold}}$ and predicted clusters $c'_j \in C_{\text{pred}}$. 
A match is identified for a gold cluster $c_i$ if there exists any predicted cluster $c'_j$ such that $c_i = c'_j$ (i.e., they contain the identical set of elements). 
Letting $N_{\text{match}}$ be the total number of unique gold clusters $c_i$ for which such a match exists, Precision is defined as $P = N_{\text{match}} / |C_{\text{pred}}|$, Recall as $R = N_{\text{match}} / |C_{\text{gold}}|$, and F\textsubscript{1} is their harmonic mean, measuring the perfect recovery of clusters.

\section{Additional experimental details}\label{sec:experimental_details}

We do not change the default temperature of the LLMs that we evaluate. 
We use the Pydantic schema for structured output (see Fig.~\ref{fig:pydantic}). 
Our system instructions for Gemini 2.0 Flash, similar to those used for other models we consider, are in Fig.~\ref{fig:gemini-prompt}. 
This represents a succinct summary of our annotation guidelines.
For our LLaMA 4 Scout experiments, we run the model in full precision on four H100 GPUs served by vLLM~\cite{Kwon2023-ej}, which takes around 3 hours to complete. 
For supervised fine-tuning with Qwen 2.5-Omni 7B, to minimize computational overhead, LoRA is applied across all layers of the model with a lightweight rank of 2, and a total training steps of 3,125 on one L40S GPU.
Each training sample is limited to 4,096 tokens to manage memory and processing time, and the total number of random samples is capped at 1,000.
Qwen 2.5-Omni 7B inference is performed with full precision with Flash Attention 2 enabled~\cite{Dao2024-wx} on four L40S GPUs, which takes around 7 hours.
For Qwen 2.5-Omni 7B LoRA SFT, we follow the instructions and configurations provided on \url{https://github.com/hiyouga/LLaMA-Factory/pull/7537}.

\begin{figure}
    \centering
\begin{lstlisting}[frame=line, basicstyle=\ttfamily\footnotesize, breaklines=true, escapeinside={(*@}{@*)}]
class ConversationalRoles(BaseModel):
    line_index: int
    reply_to: int
    speaker: str
    addressees: list[str]
    side_participants: list[str]

class ClipRoles(BaseModel):
    clip_roles: list[ConversationalRoles]    
\end{lstlisting}
    \caption{Pydantic schema.}\label{fig:pydantic}
\end{figure}

\section{Gemini error analysis}\label{sec:error_analysis}

To further illustrate these quantitative findings, we discuss an edge-case clip that models perform badly on: \textit{House M.D.}, season 4, episode 2, segment 2, clip 11:

\begin{itemize}
    \item \textit{Visual clutter}: The scene features a large number of participants, where a group of four or more doctors discusses a case in a busy hallway. This directly relates to our quantitative finding that participant count is the strongest negative predictor of performance across all conversational roles.
    \item \textit{Visual-conversational misalignment}: The editing in this scene breaks a simple alignment between sight and sound: for instance, the dialogue of the medical team is heard while the camera follows House, a different, lone character walking through another part of the hospital. Models that rely heavily on the visual presence of a speaker's face to assign roles more frequently fail here, which supports our quantitative result that lower face coverage correlates with weaker speaker attribution performance.
    \item \textit{Speech bleed-over}: The audio from one conversation begins before the speakers are visually introduced, playing over shots of other scenes or characters. This editing technique, where audio precedes the corresponding video, creates significant ambiguity that makes it difficult for a model to correctly segment and attribute the initial utterances. This aligns with our finding that lower acoustic clarity can negatively impact role attribution.
\end{itemize}

%% file: 09_annotation_guidelines.tex
\section{Annotation guidelines}\label{sec:anno_guidelines}

\paragraph{Note to the Reader.} The section outlines the conceptual framework we use to annotate conversational roles and threads in multi-party conversation.\footnote{Written primarily by AH and KC, with input and guidance from other co-authors.}
We draw on such traditions as sociolinguistics, linguistic anthropology, and television studies, especially frameworks developed by~\citet{Clark1982-am, Clark1987-pn},~\citet{Sacks1974-cg}, and~\citet{Goodwin1981-zs}, who, taken together, lay the foundation of conversation analysis and provide us with the basis of taxonomy on which our role attribution task is based; \citet{Goffman1981-fq}, on the face-to-face, social nature of dialogic interactions; and finally,~\citet{McKee2016-im}, who analyzes TV dialogues in the familiar set of vocabulary (Austinian speech acts, for instance).

Our annotation scheme is the result of several iterations, beginning with a pilot round and subsequent discussions with annotators trained in rhetoric, cognitive science, film studies, and related disciplines.
Based on observations from the pilot, we refine the definitions of utterance boundaries, reply-to structure, and role attribution to better reflect how participants---or characters on screen---engage in face-to-face dialogues in TVQA data.\footnote{This data is obtained from the original authors under TVQA Dataset Download Agreement; see \url{https://docs.google.com/forms/d/e/1FAIpQLSftzvAtUSnsmuwpKk5PU9bS7lk4FOu_v9Y2vq2nGjRXKoh6tw/viewform}}
We hope the annotation guidelines presented below will bring out the nuances of the annotation tasks at hand and their intellectual stakes, as well as provide the theoretical scaffolding that informs our design choices.

\begin{center}
    * \quad * \quad *
\end{center}

In any dialogic interaction, conversational participants manage and direct attention towards each other, during which they also inhabit various roles (speakers, addressees, etc.).
For \citet{Goffman1981-fq} in particular, conversations are more than merely linguistic exchanges but \textit{ritualized} social encounters, regulated by norms that govern such phenomena as turn-taking and floor-claiming.
These norms extend beyond linguistic structures alone and are often mediated through non-verbal and non-linguistic cues (gaze, posture, etc.).

If we follow Goffman and attempt to uncover the social architecture of interaction underlying conversational interactions, we can potentially shed light on the tacit rules and alignments that structure how people participate in talk; in the context of media studies, this enables us to probe further into issues pertaining to representation: How do characters include or \textit{exclude} others? How do they calibrate intimacy, formality, or confrontation? Who gets to speak without contest? Who must listen without reply?

We address three core questions for capturing the dynamics of conversational interactions:

\paragraph{Who initiates the conversation and controls conversational flow (holds the floor)?} This draws on McKee's notion of speech acts, as characters who initiate dialogue to \textit{perform} an action, and Goffman's idea that conversation involves the \textit{distribution of attention} among its participants. Characters who get to start a conversation and maintain the floor can be more powerful or otherwise interesting.

\paragraph{Who is explicitly addressed by the speaker?} This reflects Goodwin's and Clark's emphasis on speakers and hearers, or Goodwin's more general inquiry of conversation organization, which helps to shed light on interpersonal relationships and interactive dynamics. 

\paragraph{Who participates implicitly as side-participants or bystanders?} Building on Goffman's notion of adjacency pairs~\cite{Sacks1974-cg} and Clark's ratified vs. non-ratified participation, participants who are present but not addressed have the potential of illuminating social hierarchies, inclusion/exclusion patterns, and attentional structures within multi-party conversations. 

The annotation guidelines are organized as follows: \S\ref{sec:anno_def}: Definitions; \S\ref{sec:anno_proc}: Annotation procedure; \S\ref{sec:anno_examples}: Examples; \S\ref{sec:anno_lim}: Limitations.

\subsection{Definitions}\label{sec:anno_def}

This is grounded primarily in \citet{Goffman1981-fq,McKee2016-im}:

\paragraph{Utterance.} An utterance is a single unit of spoken dialogue, roughly equivalent to a \textit{dialogue line}. 
It is a bounded communicative act performed by a speaker within an interaction driven by the need for a plot, of the speaker, or other characters in the context of a TV show.
Unlike a \textit{statement}, which is defined by its propositional content and logical truth conditions, an utterance is defined by its timing, delivery, and interactional role.
Utterances need not be syntactically complete sentences, but as \textit{moves} that accomplish social action.
A particular type of utterance is the \textit{utterance of interest} (UOI), which is the current utterance being annotated. 
For ease of annotation, we present individual utterances as one sentence transcribed by Whisper~\cite{Radford2022-lm}. They don't have to be complete sentences, and they don't have to be a completed turn in and of itself.
This allows us to annotate conversational roles at a reasonable granularity: different lines in the same turn can have different role attributions.

\paragraph{Reply-to.} The reply-to relation connects an utterance of interest to its most relevant preceding utterance, also known as a \textit{parent} utterance, representing the line to which it responds most directly. 
This is more prevalent in NLP literature~\cite{Zhu2021-yq}, where the task is formalized as a link from a \textit{child} utterance to its \textit{parent}, and the relationship is modeled as directed graphs.
The default parent utterance is the immediately preceding line; if there is no logical antecedent utterance, the UOI is the beginning of a new thread (more below).
Since we present each utterance at the sentence level, a special form of reply-to is that of a \textit{continuation}: if the speaker is still in the midst of their turn, and the UOI merely continues the previous line, then the previous line is the natural and logical parent utterance.
In this view, we might see TV dialogues as linear sequences, where one line necessarily triggers and informs the next, until the current thread runs its course.

\paragraph{Conversational thread.} A coherent sequence of utterances sharing a sustained focus (either on a character or topic), adapted from Goffman's definition of \textit{focused interaction}. 
Dramatic scenes often involve multiple, overlapping threads, each driven by the goals and intentions of characters (aligning with McKee's concept of dramatic intent).
Its annotation involves the following steps:

\begin{itemize} 
\item Given each UOI ($u_t$), identify the single most relevant preceding utterance as its \textit{parent utterance} ($u_p$). 
\item If no suitable preceding utterance exists, annotate the UOI as replying to itself (indicating a new conversational thread). 
\end{itemize}

This annotation captures how dramatic dialogues unfold through pairs of conversational turns, marking shifts in attention, character intention, or conversational control.
Those conversational threads (also known as sub-conversations) represent the latent structure of conversations, and the central aim is to segment a broader dialogue into coherent units that represent a stable distribution of attention from conversational participants.

Crucially, our notion of a \textit{thread} is informed by Goffman's interactional order, wherein conversation is framed as a ritualized social encounter: participants are not merely exchanging information, but are engaging in a tacit choreography governed by socially shared norms of attention, responsiveness, and turn-taking.
These practices include obligations to respond when addressed (more below), to respect shifts in floor control, and monitor others' contributions for relevance, along with a variety of verbal and non-verbal cues.
As such, we emphasize \textit{both} topical \textit{and} floor change; indeed, as Goffman notes and demonstrates in his analysis, conversation is not just organized by the semantic content of the utterances.
In other words, conversational threading is fundamentally different from, e.g., running topic modeling to cluster utterances: we care about the distribution of \textit{attention} that we can observe, not that of \textit{words}.
Disentangling a conversation by correctly resolving reply-to relations serves two complementary ends: it, on the one hand, clarifies the information structure of a multi-party exchange; on the other, the observable structure of who replies to whom and when reflects the rules of engagement that make conversation intelligible and socially meaningful.

\paragraph{Conversational roles.} While disentanglement focuses on the structural linkage between utterances, conversational role attribution concerns the dynamic social positions that participants occupy throughout the course of interaction.
Each utterance is produced within a shifting \textit{ecology} of roles, an understanding of who is speaking, to whom, and in front of whom is key to interpreting both literal and, again, social meaning of an interaction.

The first role is the \textit{speaker}, the animator, the source of the utterance at time $t$. 
Following Goodwin's conversational analysis \cite{Goodwin1981-zs} and Clark's role taxonomy \cite{Clark1982-am,Clark1987-pn}, we treat conversational roles as those observably \textit{projected} at $t$: a speaker's alignment towards others (signaled through gaze, body orientation, lexical choice) casts other participants into roles.
These roles are independent of whether the projected utterance ultimately hears, understands, or replies.

\begin{table}%
\centering 
\begin{adjustbox}{max width=\linewidth}  
{
\footnotesize
\begin{tabular}{lccc}
\toprule 
\textbf{Role} & \textbf{addressed} & \textbf{ratified} & \textbf{known} \\
\midrule 
Addressee         & $+$ & $+$ & $+$ \\
Side-participants & $-$ & $+$ & $+$ \\
Bystanders        & $-$ & $-$ & $\pm$ \\
\bottomrule 
\end{tabular}}
\end{adjustbox}
\caption{Participant role matrix.} 
\label{tab:anno_participant_roles} 
\end{table}

We annotate each character present in the scene according to three dimensions, which is summarized in Table~\ref{tab:anno_participant_roles}:

\begin{enumerate} 
\item \textbf{Addressed}: Is the character explicitly spoken to by the speaker? 
\item \textbf{Ratified}: Is the character recognized by participants as legitimately part of the conversational group? 
\item \textbf{Perceptually engaged}: Is the character known (or shown visually, via multi-modal cues) to be attending to or perceiving the utterance? 
\end{enumerate}

This yields the following roles:

\begin{itemize} 
\item \textbf{Speaker}: the speaker of the line
\item \textbf{Addressee}: the \textit{intended} recipient of the line; they are directly addressed, ratified as a co-participant, and likely visibly attending to the speaker, who, in turn, likely constructs the utterance with the addressee in mind, and the addressee is expected to respond or acknowledge the utterance in some form
\item \textbf{Side-participant}: a participant who is ratified, perceptually engaged, but not explicitly addressed. This term is not really used in Goffman, but we prefer it because it has been used in previous work related to ours (e.g., \cite{Lee2011-pi}).
\item \textbf{Bystander}: a participant who is neither explicitly addressed nor ratified; may or may not be perceptually engaged (includes overhearers). If someone is not in any of the roles above, they are automatically bystanders, so we are not annotating them specifically.
\end{itemize}

\subsection{Annotation procedure} \label{sec:anno_proc}

\begin{figure*}
  \hspace*{-1.5em}
  {\centering\includegraphics[width=1.03\textwidth]{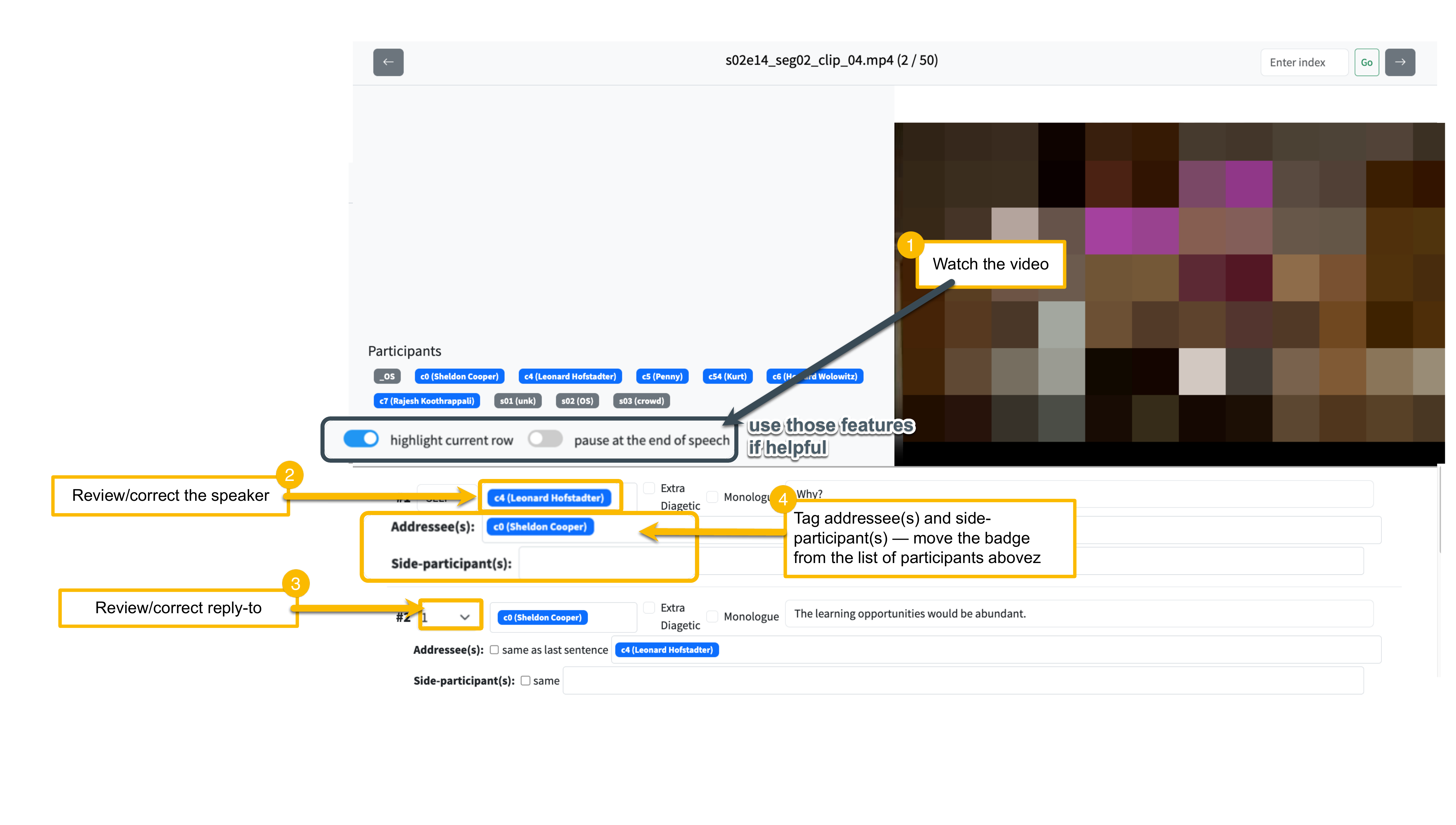}}
  
  \vspace*{-3em}
  
  \caption{Annotation interface for multimodal conversation structure understanding. Annotators proceed through four steps: a.) watch the reconstructed video clip, b.) verify or correct the speaker label, c.) review and update the reply-to link, and d.) assign addressee(s) and side-participant(s) by dragging the relevant participant badges from the list, derived from the cast list, next to the video.}\label{fig:anno1}
\end{figure*}

Annotation will take place on a dedicated interface~(Fig. \ref{fig:anno1}), and annotators will follow these concrete steps when examining each utterance in a video clip: based on the list of participants provided to you (derived from the cast list of the episode):

\begin{enumerate} 
\item Watch the reconstructed clip
\item Identify the speaker and explicitly addressed participants, which usually involves verifying or correcting existing speaker labels
\item Identify the UOI and determine its parent utterance (or mark as new thread, using~\texttt{SELF}) 
\item Identify participants who are ratified and perceptually engaged but not explicitly addressed (side-participants). Participants are represented as draggable badges, which can be moved from the original list into labeled fields
\end{enumerate}

We include the following tags to handle unidentifiable participants; they are: \textit{unknown} (not referred to by anyone else, and not identified by our face recognition pipeline, or whose identity is otherwise never revealed to the audience); \textit{crowd} (a group of unidentifiable people, such as public speaking); \textit{OS} (off screen and unknown, which presents a special edge case that we might not consider).

In annotating the clips, you are encouraged to re-watch the clip and spend as long as you require, but we recommend a time limit (of, say, 5 minutes) for each clip if this is your first pass so you will not get stuck. 
Your judgment should be entirely based on the clip and the clip alone.

\begin{figure*}
  \hspace*{-14em}
  {\centering\includegraphics[width=1.4\textwidth]{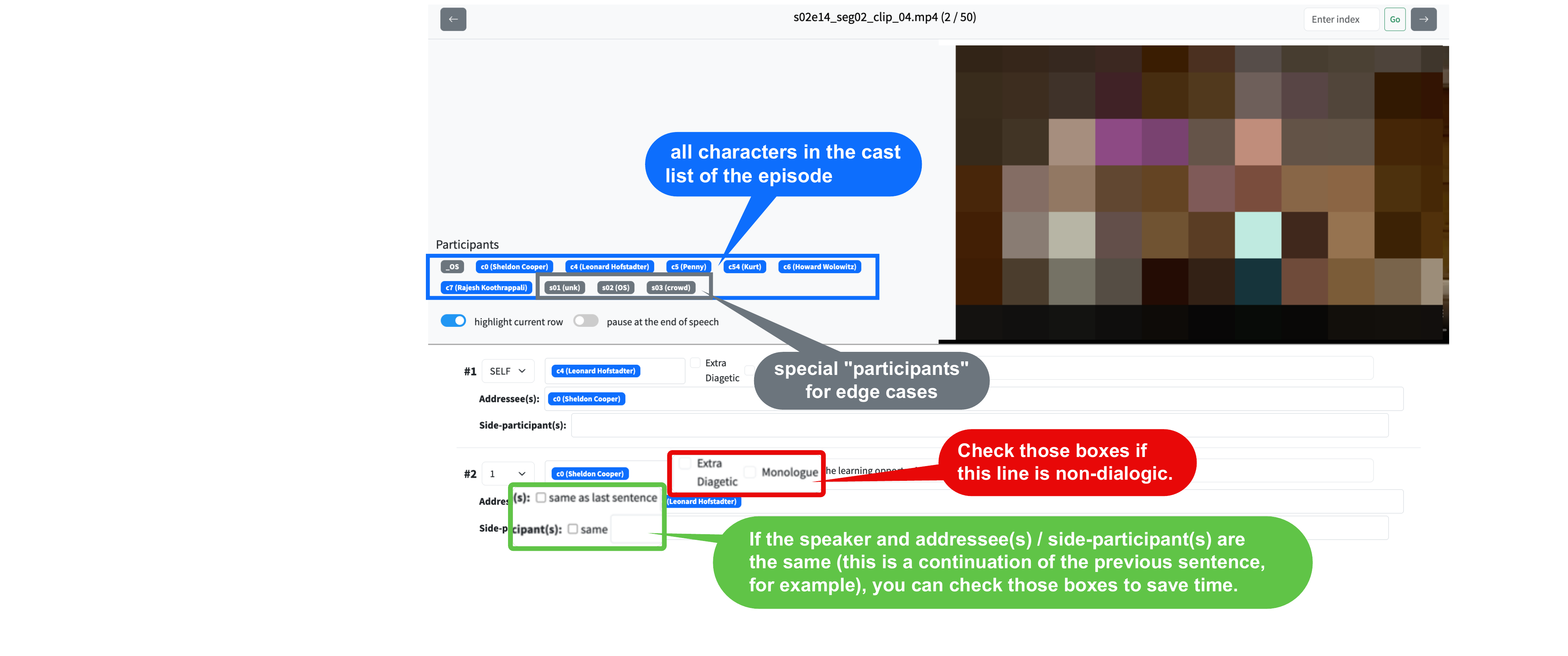}}
  \caption{Annotation enhancements for edge cases and workflow efficiency. Annotators can select characters \textit{and} special participant tokens for non-standard utterances. Checkboxes allow them to mark lines as continuations (same role as previous) or non-dialogic (extra-diegetic or monologue), with the goal to minimize redundancy and improve overall annotation consistency.}\label{fig:anno2}
\end{figure*}

\paragraph{Non-dialogic checkboxes.} Given the narrative form of the TV series, we introduce the following shorthands to indicate utterances that might not appear in typical face-to-face interactions:

\begin{itemize}
    \item \textbf{Extra-diegetic} (checkbox): to indicate narrators, ``previously on the show'', or other kinds of dialogue lines that are not part of any actual conversation
    \item \textbf{Monologue} (checkbox): to better distinguish situations where the character speaks to themselves or other inanimate objects
    \item \textbf{\texttt{\_OS} tag}: to indicate when the identity of the character is generally known to the audience but not within the scope of the clip. 
\end{itemize}

Those will give us some basis to filter out utterances that are edge cases as we assess the performance of each model on this task.
We also have a couple of other life quality features; see Fig.~\ref{fig:anno2}.

\subsection{Examples} \label{sec:anno_examples}

For a typical example, let's consider this scene taken from \textit{The Big Bang Theory} (season 2, episode 9, ``The White Asparagus Triangulation'', segment 02, clip 04).\footnote{Segment and clip number are given in TVQA.} 
In it, we observe three characters seated in a row at a movie theater, who are eligible conversational participants: Stephanie Barnett (Sara Rue), on the left, is turned slightly toward the others; Leonard Hofstadter (Johnny Galecki) sits upright, angled more towards Sheldon Cooper (Jim Parsons) on his right:

\bigskip

{\hspace*{2em}%
\begin{minipage}{\dimexpr\linewidth-2em\relax}
\begin{enumerate}[nosep]
\itemsep0em
\item[{\scshape sheldon}] I'll find us seats? \hfill (\#1) \hspace*{1em} \hphantom{i} 
\item[{\scshape stephanie}] Oh no, we have seats. \hfill (\#2) \hspace*{1em} \hphantom{i} 
\item[{\scshape leonard}] Not the right seats. \hfill (\#3) \hspace*{1em} \hphantom{i} 
\end{enumerate}
\end{minipage}}

\bigskip

Utterance \#1 opens the exchange and appears to be phrased tentatively, which suggests a proposal. Its orientation to both Stephanie and Leonard suggests a plural \textit{us} as the intended group of recipients. We don't see clearly from the visuals (no direct gaze) or vocative, so we assume both addressees are equally implicated.
Being the first utterance in the clip, this is also a thread initiator.

Stephanie provides a reply that rejects Sheldon's offer (utterance \#2); it is triggered and necessitated by utterance \#1, which is clearly its parent or reply-to utterance.
The ``we'' still invokes group reference, but the syntactic subject now reclaims authority over the seating decision.
Here, we see Stephanie looking directly at, and speaking directly to, Sheldon, which casts Leonard as a side-participant.

Leonard's utterance (\#3) is a reply to \#2, seeing as it builds on Stephanie's claim of having found seats, but disputes its adequacy. 
Semantically, we can also see this as a response to Sheldon's original proposal because it supports the need to continue looking for the seats, but structurally, we will annotate the reply-to as \#2 for two important reasons: a.) there's no reason for Leonard to say this \textit{if it wasn't for} Stephanie's utterance \#2, which makes the choice of \#1 less justifiable as it also overrides the default reply-to; b.) since threads are transitive closures of pairwise relations, and all the utterances respond to Sheldon's original need to find some seats, those three utterances will form a thread, which signals their connections, and it's important to distinguish between \textit{threads} and a single reply-to.

\subsection{Edge cases}

\paragraph{Incapacitated side-participants.} In this scene from \textit{House M. D.} (season 4, episode 2, segment 2, clip 11), a team of doctors speak about and to a patient who has locked herself into a room and is experiencing hallucinations. Although she is fully conscious, a named speaking character, and within hearing range of the conversations about her, she is not annotated as a side-participant in those conversations about her symptoms as she is preoccupied by her symptoms, not visually depicted to be listening to the conversation and is not capable of interrupting or joining the ongoing conversation. Later in the scene, however, she is labeled as an addressee after responding to her name. 

\paragraph{Off-screen side-participants.} In this scene from \textit{House M. D.} (season 4, episode 16, segment 2, clip 10), a roomful of doctors speak amongst themselves while one doctor in particular, James Wilson, sits on the side and remains largely unspeaking and unshown. Even though the camera is rarely on him and the other doctors address each other rather than him, he qualifies as a side-participant due to being in the same space and scene as the other interlocutors.

\paragraph{Non-human/inanimate objects and monologues.} In this \textit{Friends} scene (season 10, episode 06, segment 2, clip 17), the speaker addresses a duck, which appears on screen. After discussion, we concluded that addresses must be portrayed by a human in order to be counted and left the addressee blank. The label ``monologue'' distinguishes between this particular \textit{absence} of addressee from others which arise in different context (i.e., speaker talking to themselves or having no clear audience).  

\paragraph{Extra-diegetic moments.} In this scene from \textit{How I Met Your Mother} (season 6, episode 9, segment 2, clip 3), speakers cluster around a television to watch a game show, which is then shown on screen with recognizable characters and dialogue. Annotators marked all game show dialogue as extra-diegetic and did not include the TV-watchers as addressees or side-participants.

\subsection{Limitations}\label{sec:anno_lim}

The practical limitations come from our TVQA post-processing pipeline: for instance, annotators observe that cases where conversations are fast-paced (many different speakers, or many different exchanges between two interlocutors, for instance), Whisper would encounter segmentation issues: 
an utterance would be misattributed to a previous speaker's line, rather than identified accordingly as an independent utterance (as observed in \textit{House M. D.}, season 5, episode 18, segment 2, clip 20).
This error, though infrequent, unsurprisingly suggests that that clips featuring sustained, uninterrupted speech from a single speaker, whether in a series of in-scene utterances or as the result of monologic narration overlaid on a series of images,  are more likely to be correctly attributed and segmented than those with many voices, even independent of visual face counts. Future studies may focus on the role of interruption, whether verbal, visual or contextual, on multimodal conversational understandings, as well as what constitutes the boundaries of an utterance as it relates to transcription and segmentation.

\newpage

\begin{lstlisting}[frame=line]
You are a video analysis assistant.  Your task is to analyze the conversations in a video clip and its associated subtitles. For each dialogue line, you will:

*   determine what previous line it is replying to
*   determine the speaker, addressees, and side-participants

Here's how to determine the reply-to relationship between utterances to resolve conversational threads:

*   The reply-to structure gives us information about floor-claiming and topical change within the clip.
*   The character is saying this line because they want to respond to that previous line. What previous line is this current line replying to?
*   If the speaker of the last line is the same, you can treat it as continuation and put the index of last line as the reply-to.
*   If no previous line triggers this line, then write the current line index, indicating the current line replies to itself, which marks the start of a new conversational thread.

Here's how to determine each role:

*   **Speaker:**  The character who is speaking the line.  Infer this from lip movements, body language, and the context of the dialogue. If a character finishes one line and immediately starts another (very short pause), assume it's the same speaker, UNLESS there's a clear visual indication of a scene or speaker change (e.g., a camera cut to a different person starting to speak).
*   **Addressee(s):** The character(s) the speaker is *directly* addressing. Use these cues:
    *   **Eye Contact:** The most important cue. Who is the speaker looking at?
    *   **Body Orientation:** Is the speaker's body turned towards a particular person or group?
    *   **Dialogue Context:** Does the line contain a name, pronoun ("you"), or clearly refer to a specific individual or group?  ("Hey, John..." or "You all need to...")
    *   **Reactions:** If a character reacts immediately and strongly to a line (e.g., nods, responds verbally, shows surprise), they are likely an addressee.
    *   If the speaker seems to be talking to everyone present, list all characters who appear to be paying attention.
    *   If the speaker is talking to a crowd of unidentifiable characters, write "crowd".
    *   If the speaker is talking to themselves, or no one in particular, write "none".
*   **Side-Participant(s):**  Any character(s) visible in the scene *during the line's timeframe* who are *not* the speaker or addressees. They are present, and their presence is known to other participants. They can potentially join the conversation at any time.
    * If it is not possible to confidently determine if someone is a side-participant, write "unknown".
    * If there are no side-participants, write "none".

**Input:**

You will receive a list of subtitle entries.  Each entry will be a dictionary with the following keys:
*   `"line_index"`: (int) The index of the current entry (subtitle line).
*   `"start_time"`: (float) The start time of the subtitle line in seconds.
*   `"end_time"`: (float) The end time of the subtitle line in seconds.
*   `"text"`: (string) The text of the dialogue line.

You will also receive a list of potential participants for you to assign roles from. You must pick from this list.

With all this information, analyze the video segment corresponding to the `start_time` and `end_time` of each subtitle entry.

**Output:**

Provide your output in JSON format, mirroring the structure of the input.  For *each* subtitle entry, add the following keys:
*   `"line_index"`: (int) The line being analyzed.
*   `"reply_to"`: (int) The line index that this current line replies to, could be the same as the current line index or any previous line index.
*   `"speaker"`: (string) The name of the speaker. If you cannot determine the speaker, use "unknown".
*   `"addressees"`: (list of strings) A list of the names of the addressee(s).  This can be an empty list (`[]`) if there are no direct addressees, or `["none"]` if the speaker is speaking generally but to no one in particular.
*   `"side_participants"`: (list of strings) A list of the names of the side-participant(s). This can be an empty list (`[]`), `["none"]`, or `["unknown"]`
\end{lstlisting}

\captionsetup{type=figure}
\captionof{figure}{System instruction to Gemini 2.0 Flash.}\label{fig:gemini-prompt}

\newpage